\pdfoutput=1
\documentclass{article}
\usepackage[x11names]{xcolor} 
\usepackage[margin=1in]{geometry}
\usepackage{mathpazo}
\usepackage[T1]{fontenc}
\usepackage{comment}
\usepackage{bbm}
\usepackage{amsmath,amssymb}
\usepackage{todonotes}
\usepackage{pgfplots}
\pgfplotsset{compat=1.3}
\usepackage{tikz}
\usepackage{algorithm}
\usepackage{algorithmic}
\usepackage{hyperref}
\usepackage{bm}
\usepackage{booktabs}
\usepackage{longtable}
\usepackage{multicol}

\usepackage[caption=false]{subfig}

\usepackage{lettrine}
\usepackage{xspace}

\usetikzlibrary{pgfplots.groupplots}

\usepackage{color}
\usepackage[backend=biber,style=alphabetic,natbib=true,maxbibnames=99]{biblatex}
\title{Identifying Statistical Bias in Dataset Replication}
\author{Logan Engstrom\footnote{Equal contribution.} \\
        \texttt{engstrom@mit.edu} \\ 
        MIT
        \and 
        Andrew Ilyas\footnotemark[1] \\
        \texttt{ailyas@mit.edu} \\
        MIT
        \and
        Shibani Santurkar \\
        \texttt{shibani@mit.edu} \\
        MIT
        \and 
        Dimitris Tsipras  \\
        \texttt{tsipras@mit.edu} \\
        MIT
        \and 
        Jacob Steinhardt  \\
        \texttt{jsteinhardt@berkeley.edu} \\
        UC Berkeley 
        \and 
        Aleksander M\k{a}dry \\
        \texttt{madry@mit.edu} \\
        MIT
    }
\date{}

\newcommand{\va}{\texttt{v1}\xspace}
\newcommand{\vb}{\texttt{v2}\xspace}

\newcommand{\hs}[1]{\hat{s}_n(#1)\xspace}
\newcommand{\s}[1]{s(#1)\xspace}
\newcommand{\f}{f}
\newcommand{\binomial}[1]{\text{Binom}\left(n, #1\right)}
\newcommand{\dva}{\ensuremath{\mathcal{D}_1}\xspace}
\newcommand{\dvb}{\ensuremath{\mathcal{D}_2}\xspace}
\newcommand{\dvi}{\ensuremath{\mathcal{D}_i}\xspace}
\newcommand{\dvbc}{{\ensuremath{\mathcal{D}_2\vert s_1}\xspace}}

\newcommand{\dflickrc}{{\ensuremath{\mathcal{D}_{flickr}\vert s_1}\xspace}}

\newcommand{\acc}[1]{\ensuremath{\mathcal{A}_{#1}}}
\newcommand{\estacc}[1]{\ensuremath{\hat{\mathcal{A}}_{#1}}}

\newcommand{\sva}{\ensuremath{\mathcal{S}_1}\xspace}
\newcommand{\svb}{\ensuremath{\mathcal{S}_2}\xspace}

\newcommand{\pva}[1]{\ensuremath{\mathbb{P}_{x_1\sim\dva}\left(#1\right)}\xspace}

\newcommand{\eva}[1]{\ensuremath{\mathbb{E}_{x_1\sim\dva}\left[#1\right]}\xspace}
\newcommand{\evb}[1]{\ensuremath{\mathbb{E}_{x_2\sim\dvb}\left[#1\right]}\xspace}

\newcommand{\evfl}[1]{\ensuremath{\mathbb{E}_{x_{flickr}\sim\dflickr}\left[#1\right]}\xspace}

\newcommand{\naiveest}{\ensuremath{\estacc{\dvbc}^{n}}\xspace}
\newcommand{\naiveestflickr}{\ensuremath{\estacc{\dflickrc}^{n}}\xspace}
\newcommand{\dflickr}{\ensuremath{\mathcal{D}_{flickr}}}

\newcommand{\adjacc}{\acc{\dvbc}\xspace}

\newcommand{\bbn}{\text{BetaBinom}}

\DeclareMathOperator*{\argmax}{arg\,max}

\begin{document}
\maketitle
\begin{abstract}
    Dataset replication is a useful tool for assessing whether improvements in
    test accuracy on a specific benchmark correspond to improvements in models'
    ability to generalize reliably. 
    In this work, 
    we present unintuitive yet significant ways in which standard approaches to
    dataset replication introduce statistical bias, skewing the resulting
    observations.  
    We study ImageNet-v2, a replication of the ImageNet dataset
    on which models exhibit a significant (11-14\%) drop in accuracy, even
    after controlling for {\em selection frequency}, a
    human-in-the-loop measure of data quality. 
    We show that after remeasuring selection frequencies and correcting for the
    identified statistical bias, only an estimated $3.6\% \pm 1.5\%$ of the
    original $11.7\% \pm 1.0\%$ accuracy drop remains unaccounted for. 
    We conclude with concrete recommendations for recognizing and avoiding bias
    in dataset replication. 
    Code for our study is publicly
    available\footnote{\url{https://github.com/MadryLab/dataset-replication-analysis}}.
\end{abstract}

\section{Introduction}
\label{sec:introduction}
The primary objective of supervised learning is to develop models that
generalize robustly to unseen data. Benchmark test sets provide a proxy for
out-of-sample performance, but can outlive their usefulness in some cases.
For example, evaluating on benchmarks alone may steer us towards models
that adaptively overfit~\citep{reunanen2003overfitting,
bharat2008dangers,dwork2015generalization} to the finite test set and do not
generalize. Alternatively, we might select for models that are sensitive to
insignificant aspects of the dataset creation process and thus do not
generalize robustly (e.g., models that are sensitive to the exact set of
humans who annotated the test set).

To diagnose these issues, recent work has generated new, previously ``unseen''
testbeds for standard datasets through a process known as dataset replication.
Though not yet widespread in machine learning, dataset replication is a
natural analogue to experimental replication studies in the natural sciences
(cf.~\citep{bell1973experimental}). These studies play an important role in
verifying empirical findings, and ensure that results are neither affected by
adaptive data analysis, nor overly sensitive to experimental artifacts.

Recent dataset replication
studies~\citep{recht2018imagenet,recht2018cifar10,yadav2019cold} have generally
found little evidence of adaptive overfitting: progress
on the original benchmark translates to roughly the same amount (or more) of progress on
newly constructed test sets. On the other hand, model performance on the
replicated test set tends to drop significantly from the original one.

One of the most striking instances
of this accuracy drop is observed by~\citet{recht2018imagenet}, who performed a
careful replication of the ImageNet dataset and observe an 11-14\% gap between
model accuracies on ImageNet and their new test set, ImageNet-v2. The magnitude
of this gap presents an empirical mystery, and motivates us to understand 
what factors cause such a large drop in accuracy. 

In this paper, we identify a mechanism through which the dataset
replication process itself might lead to such a drop: noisy readings during data
collection can introduce statistical bias. We show that re-calibrating the
ImageNet-v2 dataset using new data while (crucially) correcting for this
bias results in an accuracy  gap of $3.6\% \pm 1.5\%$, compared to the
original $11.7\% \pm 1.0\%$ drop between ImageNet and ImageNet-v2. 

Our explanation revolves around what we refer to as the ``statistic
matching'' step of dataset replication. Statistic matching
ensures that model performance on the original test set and its
replication are comparable by controlling for variables that are
known to (or hypothesized to) impact model performance.\footnote{In causal inference
terms, statistic matching is an instance of covariate
balancing~\citep{stuart2010matching,imai2013covariate}.} Drawing a parallel to medicine, suppose we
wanted to replicate a study about the effect of a certain drug on an
age-linked disease. After gathering subjects, we have to reweight or filter
them so that the age distribution matches that of the original
study---otherwise, the results of the studies are incomparable. This
filtering/reweighting step is analogous to statistic matching in our context, 
with participant age being the relevant statistic.

To construct ImageNet-v2,~\citet{recht2018imagenet} perform statistic matching
based on the ``selection frequency" statistic, which for a given image-label pair
measures 
the rate at which crowdsourced annotators 
select the pair as correctly labeled. As we discuss in the next section,
selection frequency is a well-motivated choice of matching statistic, since
(a)~\citet{deng2009imagenet} use a similar metric to gather ImageNet images in
the first place~\citep{deng2009imagenet}, and (b)~\citet{recht2018imagenet} have
found that selection frequency is highly predictive of model accuracy.

Why does a significant drop in accuracy persist even after matching selection
frequencies? In this paper, we show that (inevitable) mean-zero noise in
selection frequency readings leads to bias in the selection frequencies of the
replicated dataset, which translates to a drop in model
accuracies. We also discuss how finite-sample reuse makes this bias difficult to
detect. 

The bias-inducing mechanism that we identify applies whenever statistic matching
is performed using noisy estimates. We characterize the mechanism theoretically in
Section~\ref{sec:id_bias}. In Section~\ref{sec:remeasure}, we remeasure
selection frequencies using Mechanical Turk and observe that as our mechanism
predicts, ImageNet-v2 images indeed have lower selection frequency on average.
After presenting a framework for studying the effect of statistical bias on
model accuracy (Section~\ref{sec:exps}), we
use de-biasing techniques to estimate a bias-corrected accuracy for
ImageNet-v2 (Section~\ref{sec:quantifying}) using the remeasured selection frequencies.
In Section~\ref{sec:implications}, we discuss the
implications of the identified mechanism for ImageNet-based computer vision
models specifically, and for data replication studies more generally.

\section{Identifying Sources of Reproduction Bias}
\label{sec:id_bias}
The goal of dataset replication is to
create a new dataset by reconstructing the pipeline that generated the original
test set as closely as possible.
We expect (and intend) for this process to
introduce a distribution shift, partly by varying parameters
that should be irrelevant to model performance (e.g. the exact identity of the
annotators used to filter the dataset). To ensure that results are
comparable with original test sets, however, dataset replication studies must
control for distribution shifts in variables that impact task performance. This
is accomplished by subsampling or reweighting the data so that each relevant
variable's distributions under the replicated dataset and the original dataset
match one another. We refer to this process as {\em statistic matching}.

Our key observation is that standard approaches to statistic matching can lead
to bias in the final replicated dataset: we illustrate this phenomenon in the 
context of the ImageNet-v2 (\vb{}) dataset
replication~\citep{recht2018imagenet}\footnote{Note that 
\citet{recht2018imagenet} actually design three datasets, in order to
measure the effect of selection frequency on model
performance. We focus our attention on \texttt{MatchedFrequency} (referred to as
just ImageNet-v2 here, in \citet{recht2018imagenet} and
elsewhere~\citep{taori2020when,li2020optimizing,ramalho2019density,ramalho2019empirical}),
since it is the only dataset designed to replicate the ImageNet validation
set.}. Before we identify the source of this bias in 
ImageNet-v2 construction,  we review the data collection process for both ImageNet
and ImageNet-v2.

\paragraph{ImageNet and selection frequency.}
ImageNet~\cite{deng2009imagenet,russakovsky2015imagenet} (which we also refer to
as ImageNet-v1 or \va{}) is one of the most widely used datasets in computer vision.
To construct ImageNet, \citet{deng2009imagenet} first amassed a large 
candidate pool of image-label pairs using image search engines such as
Flickr. The authors then asked annotators on Amazon Mechanical Turk (MTurk) to
select the candidate images that were correctly labeled.
Each image is shown to multiple annotators, and an
image's {\em selection frequency}~\footnote{Note that the term ``selection
frequency'' was in fact coined by~\citet{recht2018imagenet}, but it is also
useful for describing the initial setup of~\citet{russakovsky2015imagenet}, who
instead referred to their process as ``majority voting.''} is then defined as the
fraction of annotators that selected it.

Intuitively, images with low selection frequency are likely confusing or
their proposed label is incorrect, while images with high selection frequency should be ``easy'' for
humans to identify as the proposed label (we show examples of selection
frequencies in Figure~\ref{fig:img_vs_annotator}; further examples are in
Appendix~\ref{app:sel_freqs}). Therefore,~\citet{deng2009imagenet} include only
images with high selection frequency in the final ImageNet
dataset\footnote{Specifically, an image is included into the ImageNet test set
if a ``convincing majority''~\citep{russakovsky2015imagenet} of annotators select it.}.

\begin{figure}
\centering

\begin{multicols}{3}
\centering
\includegraphics[height=100pt]{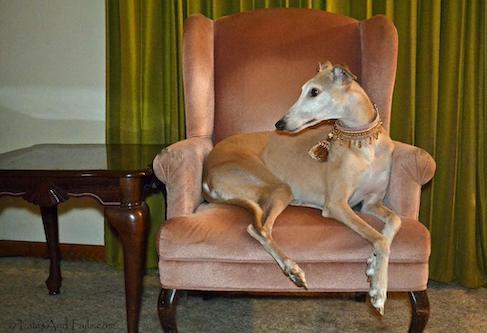}\\
Selection Frequency: 36\%
\includegraphics[height=100pt]{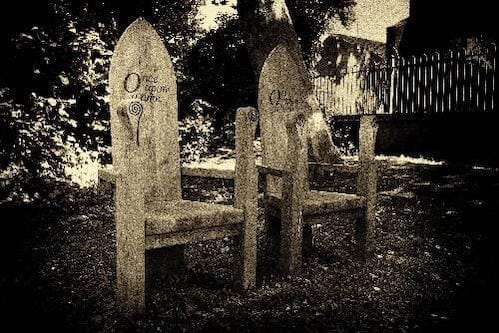}\\
Selection Frequency: 61\%
\includegraphics[height=100pt]{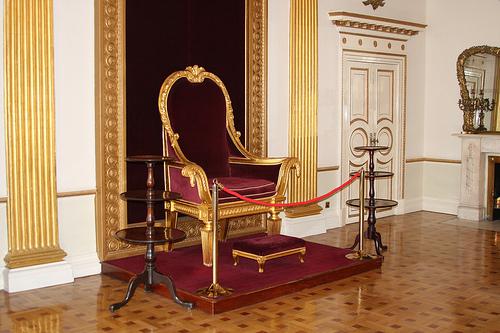}
Selection Frequency: 100\%
\end{multicols}
\caption{The smallest, median, and largest selection frequency images from
\va corresponding to the ``throne'' class (description: \textit{the chair
of state for a monarch, bishop, etc.; ``the king sat on his throne''}---the
``throne'' class was randomly chosen). The images become easier to identify
as the labeled class as selection frequency increases; for additional
context, we give a random sampling of selection frequency/image pairs in
Appendix~\ref{app:exp_setup}.}
\label{fig:img_vs_annotator}
\end{figure}

\paragraph{ImageNet-v2.}
ImageNet-v2 is a replication of ImageNet-v1 that controls for
selection frequency via statistic matching. Following the protocol
of~\citet{deng2009imagenet}, \citet{recht2018imagenet}  
collected a large pool of candidate image-label pairs, and estimated their
selection frequencies via MTurk along with a subset of the original ImageNet
validation set.~\citet{recht2018imagenet} then estimated the distribution of
ImageNet-v1 selection frequencies for each class. Finally, they subsampled ten
images of each class from the candidate pool according to the estimated
class-specific distributions.

For example, suppose 
$40\%$ of ``goldfish'' images in ImageNet-v1 have selection frequency in the
histogram bucket $[0.6, 0.8]$---when constructing ImageNet-v2,~\citet{recht2018imagenet}
would in turn sample $4$ ``goldfish'' images from the same histogram bucket in
the candidate images\footnote{Note that in~\citet{recht2018imagenet} this
process is done on a class-by-class basis and also includes provisions for when
there are not enough Flickr images in a particular bin, but the core
distribution-matching mechanism is otherwise identical.}.

Statistic matching should ensure that \va{} and \vb{} are balanced in terms of
selection frequency, and partly justifies the expectation that models should
perform similarly on both.

\begin{figure}
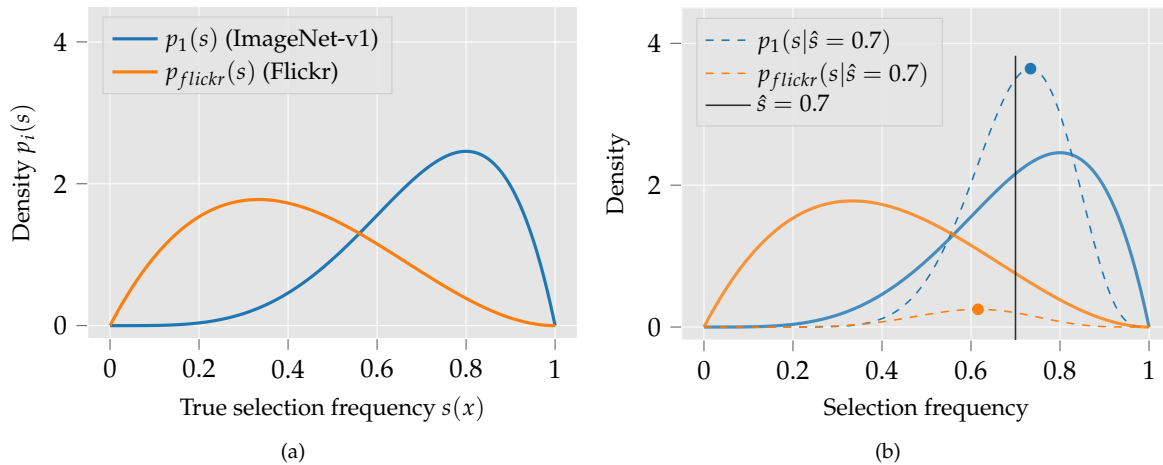

\subfloat[]{ 
\label{fig:truedists}
\input{plots/idealized_plots/trudists} 
}
\subfloat[]{
    \label{fig:s_given_s_hat}
    \input{plots/idealized_plots/cond_frequency}
}
\caption{For an image $x$, the {\em selection frequency} statistic $s(x)$
    described in
    Section~\ref{sec:id_bias} is a single number in $[0, 1]$ that captures
    how ``easy'' a given image is to classify for humans. A distribution
    over images ($p_i(x)$) thus induces a one-dimensional distribution over selection
    frequencies ($p_i(s(x))$). In {\bf (a)}, we visualize such hypothetical
    selection frequency distributions for both the Flickr data distribution
    ($p_{flickr}(s(x))$) and the ImageNet-v1 data distribution
    ($p_1(s(x))$). In {\bf (b)}, we consider a case where we are given, for
    a specific image $x$, a
    {\em noisy} version of $s(x)$ ($\hat{s}(x)$). We visualize the
    corresponding distribution of the true selection frequency $s(x)$ given this noisy 
$\hat{s}(x) = 0.7$. As discussed in Section~\ref{sec:id_bias}, note that even though
$\hat{s}(x)$ is an {\em unbiased} estimate of $s(x)$, the most likely value
of $s(x)$ for a given noisy reading of $\hat{s}(x)$ actually depends on the
distribution from which $x$ is drawn. This is the driving phenomenon behind
the observed bias between ImageNet and ImageNet-v2.}
\end{figure}

\paragraph{Sources of bias.} We identify two places where the matching strategy of~\citet{recht2018imagenet} might introduce
statistical bias. One potential source of bias could arise from binning the
images into histograms---since there are relatively few bins, it is possible
that within each bin the ImageNet images have different selection frequencies
from the corresponding Flickr images. (For example, the ImageNet-v1 images in
the $s(x) \in [0., 0.2]$ bucket might actually have selection frequency
concentrated around $0.15$, whereas the Flickr images in the same bucket might
have been concentrated around $s(x) = 0.1$.) However, this source of error
appears to have not had a pronounced effect (at least on average),
as~\citet{recht2018imagenet} report that the average selection frequency of the
ImageNet-v2 images actually matches that of the ImageNet-v1 test set.

Our analysis revolves around a second and more subtle source of
bias, however. This bias stems from the fact that for any given image $x$,
the selection frequency $s(x)$ is never
measured exactly. Instead, we are only able to measure $\hat{s}(x)$, a
finite-sample estimate of the statistic, attained by averaging over a
relatively small number of annotators.

To model the impact of this seemingly innocuous detail, suppose that the
selection frequencies $s(x)$ of ImageNet and Flickr images are distributed
according to $p_1(s(x))$ and $p_{flickr}(s(x))$ respectively (or more briefly,
$p_1(s)$ and $p_{flickr}(s)$)---see
Figure~\ref{fig:truedists} for a visualization. Now, suppose that for an image $x$, we get an
unbiased noisy measurement $\hat{s}(x) = 0.75$ of the selection frequency via
crowdsourcing. Then, even if $\hat{s}(x)$ is an unbiased estimate of $s(x)$, the
most likely value of $s(x)$ for the image is not $\hat{s}(x)$, but in fact
depends on the distribution from which $x$ was drawn. Indeed, for the
(hypothetical) distributions shown in Figure~\ref{fig:truedists}, if $x$ is a Flickr image then
it is more likely that $s(x) < 0.75$ and $\hat{s}$ is an overestimate, since a
priori an image is likely to have a low selection frequency (i.e.,
there is more $p_{flickr}(s)$ mass below $0.75$) and the noise is unbiased.
Conversely, if $x$ is an ImageNet test set image in this same setting, it is
more likely that $s(x) > 0.75$. Therefore, if we use a Flickr image with a noisy
selection frequency $0.75$ to ``match'' an ImageNet image with the same noisy
selection frequency, the true selection frequency of the ImageNet image is
actually likely to be higher. We can make this explicit by writing down the
likelihood of $s$ given $\hat{s} = 0.75$ (also plotted in
Figure~\ref{fig:s_given_s_hat}):
$$p_i(s|\hat{s}=0.75) = \frac{p_i(s) \cdot
p(\hat{s}=0.75|s)}{p_i(\hat{s}=0.75)} \qquad\forall\ i \in \{1,\ flickr\},$$
which depends on the prior $p_i(\cdot)$ and therefore is not equal for both
values of $i$.

The distribution of candidate Flickr images is likely skewed to have lower
selection frequencies than \va---after all,
\citet{deng2009imagenet} narrowed down Imagenet-v1 from a large set of candidates based on quality.
Therefore, one would expect the underlying true selection frequencies of
the \va{} images to be higher than their matched ImageNet-v2 counterparts (as
illustrated in Figure ~\ref{fig:truedists}). More generally, bias arises from
the fact that even though $\hat{s}(x)$ given $x$ is unbiased (e.g.
$E[\hat{s}(x)\vert x] = E[\hat{s}(x)\vert s(x)] = s(x)$), functions of $x$ given
$\hat{s}(x)$ are not necessarily unbiased for many natural statistics (i.e. $E[q(x)\vert \hat{s}(x)] \neq
E[q(x)\vert s(x)]$ for statistics $q$).

\begin{figure}
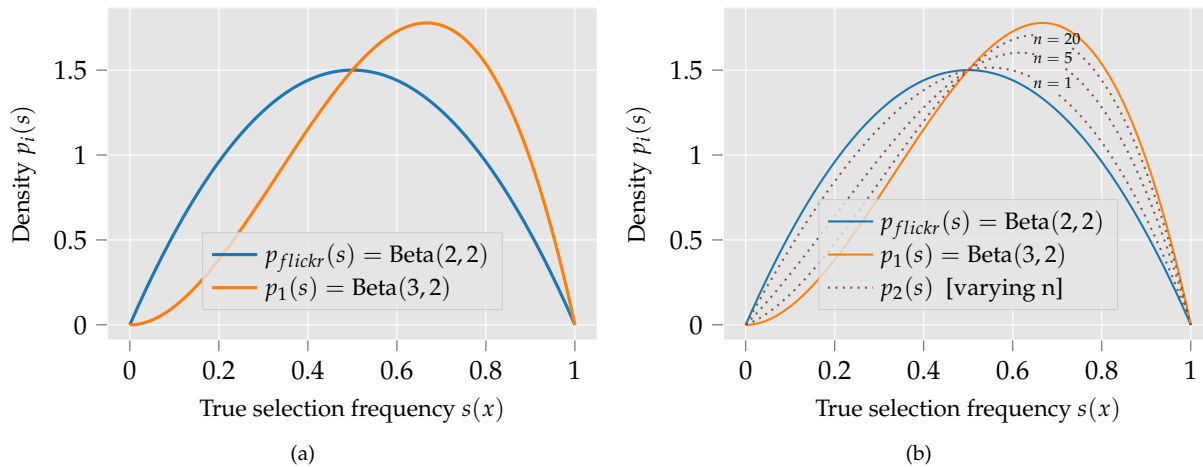

\subfloat[]{ 
    \label{fig:theory_1}
    \input{plots/idealized_plots/theory_dists} 
}
\subfloat[]{
    \label{fig:theory_2}
    \input{plots/idealized_plots/theory_results}
}
\caption{Illustrations accompanying the simple theoretical model. {\bf (a)} In
    the simple model, we assume $p_1(s(x))$ and $p_{flickr}(s(x))$ are
$\text{Beta}(\alpha+1,\beta)$ and $\text{Beta}(\alpha,\beta)$,
respectively---this is visualized above for the case of $\alpha=\beta=2$. {\bf
(b)} The results of the simple model reveal that as more and more samples are used
to estimate $s(x)$ for each image, the resulting ImageNet-v2 distribution tends
towards the \va{} distribution, but does not actually match the \va{} sample for
any finite number of samples per image.}
\end{figure}

\paragraph{A simple model of the bias.} To better understand the
source of the bias, consider a simple model in which the ImageNet-v2 selection
process is cast as a rejection sampling procedure. Here, the densities
$p_{1}(\hat{s}(x))$ and $p_{flickr}(\hat{s}(x))$ are estimated from samples
(analogous to the histograms of~\citet{recht2018imagenet})---then, for a given
Flickr image $x$, we ``accept'' $x$ into the \vb{} dataset with probability
proportional to $p_{1}(\hat{s}(x)) / p_{flickr}(\hat{s}(x))$ (analogous to the
bin-wise sampling of~\citet{recht2018imagenet}). If selection
frequency readings were not noisy, i.e. if $\hat{s}(x) = s(x)$, then the
resulting density of selection frequencies in the \vb{} dataset would be given
by
$$p_{flickr}(s(x)) \cdot \frac{p_{1}(s(x))}{p_{flickr}(s(x))} =
p_{1}(s(x)),$$
and the selection frequencies of \vb{} would be distributed in
the same way as those of \va{}, as intended. However, the inevitable
noisiness of the selection frequencies means that in reality, the density
of selection frequencies for \vb{} will be given by
\begin{align*}
p_{flickr}(s(x)) \cdot \mathbb{P}(\text{x is accepted into \vb}\vert s(x)) &=
p_{flickr}(s(x)) \cdot \int_{\hat{s}} p(\hat{s}\vert s) \mathbb{P}(\text{x
is accepted} \vert \hat{s}(x)) \\
&= p_{flickr}(s(x)) \cdot \int_{\hat{s}} p(\hat{s}\vert s)
\frac{p_{1}(\hat{s}(x))}{p_{flickr}(\hat{s}(x))}.
\end{align*}
Now, suppose for simplicity that $p_{flickr}(s)$ and $p_{1}(s)$ are given by beta
distributions $\text{Beta}(\alpha, \beta)$ and $\text{Beta}(\alpha+1,
\beta)$ respectively\footnote{This is a simple
model intended for illustrative purpose---we will later use a more sophisticated
model to capture the actual distribution of selection frequencies.} (c.f.
Figure~\ref{fig:theory_1}). Furthermore, suppose that $\hat{s}(x)$ is given by
an average of $n$ Bernoulli draws with success probability $s(x)$. Then, a
series of calculations (shown in Appendix~\ref{app:theory_model}) reveals that
the resulting \vb{} selection frequency distribution is given by:
\begin{equation}
\label{eq:toymodelsoln}
\frac{n}{n+\beta+\alpha}\cdot \text{Beta}(\alpha+1, \beta) +
\frac{\alpha+\beta}{n+\beta+\alpha}\cdot \text{Beta}(\alpha,\beta) =
\frac{n}{n+\beta+\alpha}\cdot p_{1}(s) +
\frac{\alpha+\beta}{n+\alpha+\beta} \cdot p_{flickr}(s).
\end{equation}
Note that as $n \rightarrow 0$ (no filtering is done at all), the above expression
evaluates to exactly $p_{flickr}(s)$, as expected. Then, as the number of
workers $n$ tends to infinity (i.e. $\hat{s}$ becomes less noisy), the
distribution of ImageNet-v2 selection frequencies
converges to the desired $p_{1}(s)$. For any finite $n$,
however, the resulting \vb{} distribution will be a non-degenerate mixture
between $p_{flickr}(s)$ and $p_{1}(s)$, and therefore does not match the
distribution of selection frequencies $p_{1}(s)$ exactly. The results of
this toy model (depicted in Figure~\ref{fig:theory_2}) capture
the bias that could be incurred by the data replication pipeline
of~\citet{recht2018imagenet}. In the next section, we set out to
quantify the bias suffered by the actual pipeline
of~\citet{recht2018imagenet}.

\section{Remeasuring Selection Frequencies}
\label{sec:remeasure}

In this section, we measure the effect of the described noise-induced bias on
the true and observed selection frequencies of images in \v and \vb.
Using an annotation task closely resembling those of the ImageNet-v2 and
ImageNet MTurk experiments, we collect new selection frequency estimates for all
of ImageNet-v2 and for a subset of ImageNet. In these tasks, MTurk annotators were
shown grids of 48 images at a time, each corresponding to an ImageNet
class. Each grid contained a mixture of ImageNet, Flickr, and in our case,
ImageNet-v2 images of the corresponding class (since
ImageNet-v2 was not yet realized at the time of the other experiments), as well
as control images from other classes. We describe the setup in more detail in 
Appendix~\ref{app:remeasure_setup}. Annotators were 
tasked with selecting all the images in the grid containing an object from the
class in question. Each image was seen by 40 distinct annotators, and assigned an
observed selection frequency equal to the fraction of these workers that
selected it.

\begin{figure}
    \centering \input{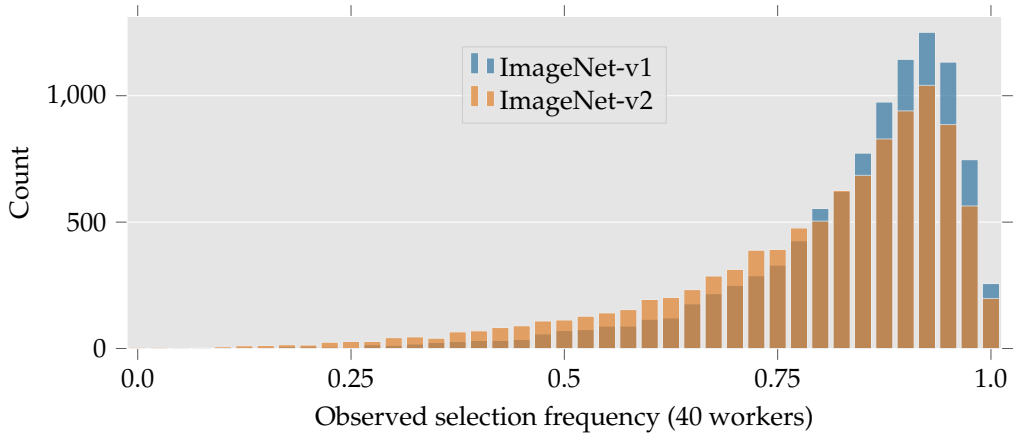} \caption{Selection frequency
    histograms for \va~and \vb based on our selection frequency re-measurement
    experiment. Results indicate that \vb~seems to have lower 
    selection frequency.} \label{fig:histograms}
\end{figure}

Histograms of observed selection frequencies for \va~and \vb~are shown in
Figure~\ref{fig:histograms}. We find that the average selection frequencies of
the \va~and \vb~images were {85.2\%} $\pm$ 0.1\%~and {80.7\%} $\pm$ 0.1\%
respectively compared to {71\%} and {73\%} reported by
\citet{recht2018imagenet}~\footnote{95\% bootstrapped CI. The average selection frequency (unlike 
other statistics we discuss) is always unbiased, and thus average observed
selection frequency will converge quickly to the average true selection
frequency by law of large numbers.}. This means that the initial 2\% gain in
selection frequency measured by~\citet{recht2018imagenet} turns into a 5\%
drop\footnote{Our model in Section~\ref{sec:id_bias} predicts a distributional
difference in selection frequencies between \va and \vb; a gap between means is
sufficient but not necessary evidence for this difference.}. Our model of dataset
replication bias predicts this discrepancy: once observed selection frequencies
are used for statistic matching, they no longer provide an unbiased estimate of
true selection frequency\footnote{To draw an analogy here, suppose that instead
of matching image datasets, we were matching two piles of coins: a rigged pile A
($P_A(\text{heads})=1$) and a fair pile B ($P_B(\text{heads}) = 0.5$). To match
pile B to pile A, we flip the coins in both piles 10 times each---inevitably,
some of the coins in pile B will land heads all 10 times. These ``matched''
coins are identical to pile A coins with respect to the observed ``number of
heads'' statistic, but are obviously not identical coins. Yet, even though
flipping the matched coins another 10 times would reveal this, it is impossible
to conclude anything other than $P(\text{heads}) = 1$ solely from the
already-collected data on the matched coins.}.

\paragraph{Detecting bias using the original data. }
Our MTurk task measures a significant selection frequency gap betweeen
\va and \vb (\textasciitilde 5\%), but also measures average selection frequencies for both
datasets to be significantly higher than reported
by~\citet{recht2018imagenet}, suggesting differences in experimental setup.
Indeed, while the tasks themselves were identical, we did make a few changes to
the deployment setup of~\citet{recht2018imagenet} to improve data quality. These
changes are outlined in Appendix~\ref{app:exp_differences}: examples include
introducing worker screening qualifications\footnote{Worker qualification is a
service provided by MTurk that only allows ``high-reputation'' annotators 
(typically measured by historical annotation quality on the platform) to
complete a given task. Qualifications have been shown to significantly impact data
quality: in~\citep{peer2013reputation}, using qualifications lowered the
number of inattentive workers from 33\% to less than 1\%.}, and using different
proportions of images per grid. Since the task interface remained constant and workers are not
able to distinguish between ImageNet-v1 and ImageNet-v2 images while labeling,
we believe that the changes made improve data quality across both datasets while
negligibly affecting the selection frequency gap between them.

Still, we can fully control for experimental differences by analyzing the raw
data of~\citet{recht2018imagenet} directly, taking care to avoid bias from observed
selection frequency reuse.
We defer the exact data analysis to Appendix~\ref{app:orig_data}; in summary, we
perform three experiments:
\begin{itemize}
    \item Recall that~\citet{recht2018imagenet} perform statistic matching using
    observed selection frequencies, measured with $n=10$ annotators, to get
    ImageNet-v2. We gradually decreased $n$ to study the effect of finite-sample
    noise, and found that model accuracy on the resulting replicated dataset
    degrades. For example, the accuracy gap from \va to the replication
    increases from 12\% when $n = 10$, to 14\% when $n = 5$. 
    This is consistent with our model of statistic matching bias: fewer
    annotators means noisier observed selection frequencies $\hs{x}$, which in
    turn amplifies the effect of the bias, driving down model accuracies.

    \item In the second experiment, we repeat the statistic matching process
    of~\citet{recht2018imagenet} using 
    five ``in-sample'' annotators per candidate image, reserving the remaining
    annotations as a held-out set. 
    We find that the average selection frequency measured by the
    in-sample annotations overestimates the true average selection frequency
    (i.e., as measured by the held-out set) by 2-3\%. 

    \item Finally, we use the held-out set from this second experiment to
    filter the candidate pool via a held-out selection frequency cutoff. 
    This skews the distribution of true selection frequencies in the candidate
    pool towards higher values.
    As predicted by our model of statistic-matching bias, using this skewed
    subset in place of the full candidate pool for statistic matching results in
    increased model accuracy (yet identical in-sample selection frequencies). 
\end{itemize}

These results suggest that statistic matching bias affects the \vb dataset, even
fully controlling for experimental setup. In the coming sections, we
quantify the effects of this bias on the model accuracies observed
by~\citet{recht2018imagenet}.

\section{Understanding the Accuracy Gap}
\label{sec:exps}
Our findings so far have suggested that statistic matching bias results in a
downwards bias in ImageNet-v2 true selection frequencies. 
In this section, we quantify the impact on this bias on ImageNet-v2 accuracy. 

\subsection{Notation and terminology}
Here we overview the notation and terminology useful in discussing the bias in
ImageNet-v2 accuracy. 

\paragraph{Selection frequencies.} In Section~\ref{sec:id_bias} we defined the
true selection frequency $s(x)$ for an image $x$ to be the (population) rate at
which crowd annotators select the image as ``correctly labeled.'' The true
selection frequency of an image is unobservable, and often approximated by the
observed selection frequency, $\hs{x} \sim \frac{1}{n} \binomial{\s{x}}$, which
can be estimated from an $n$-annotator MTurk experiment.
When $n$ is clear from context we will often omit it and write $\hat{s}(x)$.

\paragraph{Distributions.} We will use $\dva$ and $\dvb$ to denote the
distributions of \va and \vb images respectively,
and $\sva$ and $\svb$ to denote the corresponding finite test sets.
As in Section~\ref{sec:id_bias}, we
denote by $p_i(s)$ the probability density of true selection frequencies for
images drawn from $\dvi$ (or Flickr, if $i=flickr$). Similarly, we use
$p_i(\hs{x})$ to denote the probability mass function of the
observed selection frequency for dataset $i$.

We let $\dvbc$ be the distribution of ImageNet-v2 images reweighted to have the
same selection frequency distribution as ImageNet-v1. Formally, $\dvbc$ is
the compound distribution $\left( x_2 \sim \dvb \vert s(x_2) 
\sim p_1(s) \right)$. Sampling from $\dvbc$ corresponds to first sampling a
\va{} image $x_1$, then sampling an image $x_2$ from the \vb{} distribution
conditioned on $s(x_2) = s(x_1)$.  

\paragraph{Accuracies.} For a classifier $c$, let $\f_c(x)$ be an indicator
variable of whether $c$ correctly classifies $x$. Since our analysis applies to
any fixed classifier $c$, we omit it and use $\f(x)$.
We then define $\acc{X}$ to represent classifier accuracy on distribution or
test set $X$---for example, classifier accuracy on \va{} is given by
$$\acc{\dva} = \pva{\f(x_1) = 1} = \eva{\f(x_1)}.$$ 

\subsection{Breaking down the accuracy gap} 
The accuracy gap between the \va{} and \vb{} test sets is given by $\acc{\sva} -
\acc{\svb}$. What fraction of this gap can be attributed to bias in selection
frequency? To answer this, we decompose this accuracy gap into three elements
whose contribution can be studied separately:
\begin{align}
  \label{eq:decomp}
    \acc{\sva} - \acc{\svb}  
    &= \underbrace{\left(\acc{\sva} - \adjacc\right)}_{\text{bias-corrected accuracy gap}}
            + \underbrace{\left(\adjacc - \acc{\dvb}\right)}_{\text{selection gap}} +
            \underbrace{\Big(\acc{\dvb} - \acc{\svb}\Big)}_{\text{finite
            sample gap}\ \approx\ 0}.
\end{align}

\paragraph{Bias-corrected accuracy gap.} The first term of~\eqref{eq:decomp},
called the bias-corrected accuracy gap, captures the portion of the \va-\vb
accuracy drop that {\em cannot} be explained by a difference in selection
frequency, and instead might be explained by benign distribution shift or
adaptive overfitting\footnote{In fact, this term can be further decomposed into
a sum of an adaptivity gap $(\acc{\sva} - \acc{\dva})$, and a distribution shift
gap $(\acc{\dva} - \adjacc).$ However, since we don't have access to
$\acc{\dva}$ it is difficult to disentangle these.}.

\paragraph{Selection gap.} The second term of~\eqref{eq:decomp} is accuracy gap
that can {\em only} be attributed to selection frequency, since it
compares accuracy on $\dvb$ to accuracy on a reweighted version of $\dvb$.
If there was no bias, and the distribution of selection frequencies for \va{}
and \vb{} matched exactly, then this term would equal zero ($\dvbc$ would equal
$\dvb$).  
Thus, the selection gap translates the effect of discrepancy in true selection
frequency between \va and \vb into a discrepancy in accuracy.
Since we measured \va as having higher true selection frequency,  we expect the
selection gap to be positive and thus explain a portion of the accuracy gap that
was previously attributed to distribution shift.

\paragraph{Finite-sample error.} The final term refers to the
finite-sample error from using $10,000$ images as a proxy for
distributional accuracy. We believe that this term is negligible, since (a) 95\%
bootstrapped confidence intervals for the classifiers we evaluate are all at
most $0.1\%$, and (b) there can be no adaptive overfitting on $\svb$ with
respect to $\dvb$. Thus, we drop this term from consideration and instead use
$\acc{\dvb}$ and $\acc{\svb}$ interchangeably.\\

\paragraph{Computing selection-adjusted accuracy.} We have shown how to
decompose the \va-\vb accuracy gap into a component explained by selection
frequency (selection gap), and a component unexplained by selection frequency
(bias-corrected accuracy gap). 
The challenge in computing this decomposition is estimating $\adjacc$, the
selection-adjusted \vb{} accuracy. While the closed form of $\adjacc$ is
\begin{align*}
    \int_s \mathbb{E}_{\dvb}[f(x)|s(x)=s] \cdot p_1(s)\ ds,
\end{align*}
we have no access to $p_i(s)$ for any value of $i$ (we do not even have direct
access to $s(x)$ for any image $x$). In the next section, we explore methods for
estimating $\adjacc$ using only the observed selection frequencies that we
collected.

\section{Quantifying the Bias}
\label{sec:quantifying}
In the previous sections, we showed that statistic matching based on noisy
observed selection frequencies may lead ImageNet-v2 images to have lower true
selection frequencies than expected. 
In Section~\ref{sec:exps} we related this discrepancy in selection frequency to
a corresponding discrepancy in model accuracy between \va and \vb, which we
called the ``selection gap.'' 
In this section, we explore a series of methods for estimating this gap---we
estimate that the selection gap accounts for 8.1\% of the 11.7\% \va{}-\vb{}
accuracy drop. 

\subsection{Na\"{i}ve approach}
\label{sec:naive}
In the last section we introduced the selection-adjusted \vb accuracy,
\begin{equation}
    \label{eq:adjacc_true}
    \adjacc = \int_s \mathbb{E}_{\dvb}[f(x)|s(x)=s] \cdot p_1(s)\ ds,
\end{equation}
which captures model accuracy on a version of ImageNet-v2 reweighted to have the
same true selection frequency distribution of ImageNet-v1. Since we do not
observe true selection frequencies, we cannot evaluate $\adjacc$, and are
instead forced to estimate it. A natural way to do so 
is to use observed selection frequencies in place of true ones, leading
to the following ``na\"ive estimator:''
\begin{align}
    \label{eq:adjacc_hat} 
    \naiveest = \sum_{k=0}^{n} \evb{\f(x_2)\vert \hs{x_2} = \frac{k}{n}} \cdot p_1\left(\hs{x_1} =\frac{k}{n}\right).
\end{align}
The na\"ive estimator is a computable\footnote{We can compute the na\"ive
estimator as long as we have enough images to reliably approximate the
expectations. We assume this is the case in our study, since (a) we have $10^4$
images and only $41$ possible values of $\hs{x}$ and (b) halving the number of images
negligibly affects the value of the estimator.} but biased estimator of the
selection-adjusted accuracy. This follows from our analysis in
Section~\ref{sec:id_bias}, since $\naiveest$ is just a mechanism for statistic
matching between ImageNet-v1 and ImageNet-v2 using observed selection
frequencies in place of true selection frequencies. 
Thus, the selection-adjusted \vb{} accuracy computed by the na\"ive estimator
is likely to still underestimate the true selection-adjusted accuracy
$\adjacc$.

\begin{figure}
\subfloat[]{
\label{subfig:ps_workers_raw}
\begin{tikzpicture}

\definecolor{color0}{rgb}{0.83921568627451,0.152941176470588,0.156862745098039}
\definecolor{color1}{rgb}{0.172549019607843,0.627450980392157,0.172549019607843}
\definecolor{color2}{rgb}{1,0.498039215686275,0.0549019607843137}
\definecolor{color3}{rgb}{0.12156862745098,0.466666666666667,0.705882352941177}

\begin{axis}[
width=0.33\linewidth,
axis background/.style={fill=white!89.8039215686275!black},
axis line style={white},
legend cell align={left},
legend style={fill opacity=0.8, draw opacity=1, text opacity=1, at={(0.03,0.97)}, anchor=north west, draw=white!80!black, fill=white!89.8039215686275!black},
tick align=outside,
tick pos=left,
x grid style={white},
xlabel={\small Observed selection frequency},
xmajorgrids,
xmin=-0.05, xmax=1.05,
xtick style={color=white!33.3333333333333!black},
y grid style={white},
ylabel={\small Estimated density \(\displaystyle p_1(s)\)},
ymajorgrids,
ymin=-0.197, ymax=4.313,
ytick style={color=white!33.3333333333333!black}
]
\addplot [very thick, color0, opacity=1.0, mark=*, mark size=1, mark options={solid}]
table {%
0 0.02
0.025 0.016
0.05 0.016
0.075 0.008
0.1 0.028
0.125 0.044
0.15 0.048
0.175 0.052
0.2 0.052
0.225 0.08
0.25 0.128
0.275 0.104
0.3 0.168
0.325 0.18
0.35 0.148
0.375 0.2
0.4 0.268
0.425 0.32
0.45 0.344
0.475 0.468
0.5 0.432
0.525 0.436
0.55 0.568
0.575 0.604
0.6 0.74
0.625 0.832
0.65 0.796
0.675 1.028
0.7 1.216
0.725 1.44
0.75 1.568
0.775 1.764
0.8 1.888
0.825 2.344
0.85 2.628
0.875 3.056
0.9 3.704
0.925 4.108
0.95 3.92
0.975 3.008
1 1.228
};
\addlegendentry{\scriptsize 40 annotators}
\addplot [very thick, color1, opacity=1.0, mark=*, mark size=1, mark options={solid}]
table {%
0 0.016
0.05 0.028
0.1 0.026
0.15 0.052
0.2 0.094
0.25 0.124
0.3 0.164
0.35 0.198
0.4 0.282
0.45 0.34
0.5 0.49
0.55 0.6
0.6 0.726
0.65 0.928
0.7 1.194
0.75 1.462
0.8 1.89
0.85 2.548
0.9 3.234
0.95 3.418
1 2.186
};
\addlegendentry{\scriptsize 20 annotators}
\addplot [very thick, color2, opacity=1.0, mark=*, mark size=1, mark options={solid}]
table {%
0 0.024
0.1 0.05
0.2 0.106
0.3 0.187
0.4 0.312
0.5 0.489
0.6 0.756
0.7 1.162
0.8 1.824
0.9 2.513
1 2.577
};
\addlegendentry{\scriptsize 10 annotators}
\addplot [very thick, color3, opacity=1.0, mark=*, mark size=1, mark options={solid}]
table {%
0 0.047
0.2 0.1525
0.4 0.3705
0.6 0.7325
0.8 1.5175
1 2.18
};
\addlegendentry{\scriptsize 5 annotators}
\end{axis}

\end{tikzpicture}}
\subfloat[]{
\label{subfig:p_f_given_s}
\begin{tikzpicture}

\definecolor{color0}{rgb}{0.83921568627451,0.152941176470588,0.156862745098039}
\definecolor{color1}{rgb}{0.172549019607843,0.627450980392157,0.172549019607843}
\definecolor{color2}{rgb}{1,0.498039215686275,0.0549019607843137}
\definecolor{color3}{rgb}{0.12156862745098,0.466666666666667,0.705882352941177}

\begin{axis}[
width=0.33\linewidth,
axis background/.style={fill=white!89.8039215686275!black},
axis line style={white},
tick align=outside,
tick pos=left,
x grid style={white},
xlabel={\small Observed selection frequency},
xmajorgrids,
xmin=-0.05, xmax=1.05,
xtick style={color=white!33.3333333333333!black},
y grid style={white},
ylabel={\small ResNet-26 accuracy},
ymajorgrids,
ymin=-0.045928338762215, ymax=0.964495114006515,
ytick style={color=white!33.3333333333333!black}
]
\addplot [very thick, color0, opacity=1.0, mark=*, mark size=1, mark options={solid}]
table {%
0 0
0.025 0
0.05 0
0.075 0
0.1 0
0.125 0.0909090909090909
0.15 0.0833333333333333
0.175 0.0769230769230769
0.2 0.153846153846154
0.225 0.05
0.25 0.15625
0.275 0.153846153846154
0.3 0.0952380952380952
0.325 0.2
0.35 0.108108108108108
0.375 0.04
0.4 0.164179104477612
0.425 0.1
0.45 0.13953488372093
0.475 0.188034188034188
0.5 0.240740740740741
0.525 0.247706422018349
0.55 0.330985915492958
0.575 0.304635761589404
0.6 0.302702702702703
0.625 0.307692307692308
0.65 0.381909547738693
0.675 0.404669260700389
0.7 0.447368421052632
0.725 0.444444444444444
0.75 0.510204081632653
0.775 0.562358276643991
0.8 0.597457627118644
0.825 0.627986348122867
0.85 0.694063926940639
0.875 0.74738219895288
0.9 0.779697624190065
0.925 0.831548198636806
0.95 0.860204081632653
0.975 0.894946808510638
1 0.9185667752443
};
\addplot [very thick, color1, opacity=1.0, mark=*, mark size=1, mark options={solid}]
table {%
0 0
0.05 0
0.1 0
0.15 0.135135135135135
0.2 0.111111111111111
0.25 0.0983606557377049
0.3 0.146341463414634
0.35 0.166666666666667
0.4 0.156716417910448
0.45 0.225130890052356
0.5 0.26431718061674
0.55 0.341637010676157
0.6 0.316076294277929
0.65 0.401315789473684
0.7 0.481099656357388
0.75 0.506443298969072
0.8 0.624242424242424
0.85 0.704600484261501
0.9 0.757802746566791
0.95 0.84134337000579
1 0.869646182495345
};
\addplot [very thick, color2, opacity=1.0, mark=*, mark size=1, mark options={solid}]
table {%
0 0.0434782608695652
0.1 0.0819672131147541
0.2 0.164835164835165
0.3 0.192708333333333
0.4 0.205438066465257
0.5 0.305283757338552
0.6 0.361413043478261
0.7 0.522514868309261
0.8 0.641456582633053
0.9 0.74333855799373
1 0.835497835497836
};
\addplot [very thick, color3, opacity=1.0, mark=*, mark size=1, mark options={solid}]
table {%
0 0.133928571428571
0.2 0.20440251572327
0.4 0.349397590361446
0.6 0.471100062150404
0.8 0.641876046901173
1 0.775046382189239
};
\end{axis}

\end{tikzpicture}}
\subfloat[]{
\label{subfig:adj_acc_workers}
\input{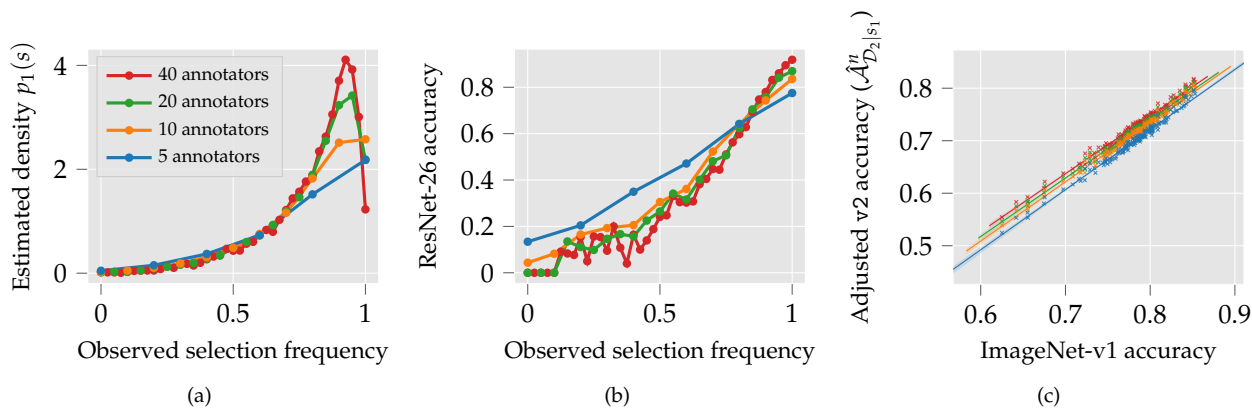}}
    \caption{A series of graphs, all demonstrating
        bias in estimators that condition on selection frequency.
        {\bf Left}: The estimated population density of selection frequencies,
        calculated na\"ively from samples. For a given number
        of annotators per image $n$, the corresponding line in the graph
        has equally spaced points of the form $(k/n, \sum
        \bm{1}_{\hat{s}=k/n})$. {\bf Middle}: Model accuracy
        of a ResNet-26 conditioned on selection frequency; once again, we
        na\"ively using empirical selection frequency in place of true selection
        frequency for conditioning. Just as in the left-most graph, for a given
        $n$-annotator line, points at $x=k/n$ 
        in the graph correspond to the accuracy on images with observed
        selection frequency $k/n$. {\bf Right}: Adjusted \va~versus
        \vb~accuracy plots, calculated for varying numbers of annotators
    per image (with adjusted accuracy computed using the na\"ive estimator of
    Section~\ref{sec:naive}). Each point in the plot corresponds to a trained model.}
       \label{fig:showbias}
\end{figure}

We can verify this bias empirically by varying the number of annotators $n$ used
to calculate $\hat{s}_n(x)$ for each image, and visualizing the resulting trends in
$p_i(\hs{x})$ (Figure~\ref{subfig:ps_workers_raw}), $p_i(\f(x) = 1\vert \s{x})$
(Figure~\ref{subfig:p_f_given_s}), and $\naiveest$
(Figure~\ref{subfig:adj_acc_workers}). The results corroborate our analysis
in Section~\ref{sec:id_bias} and our findings from
Section~\ref{sec:remeasure}. Specifically, Figure~\ref{fig:showbias} plots each
term in the definition of the na\"ive estimator,
\begin{align}
    \underbrace{\naiveest}_{\text{Fig. \ref{subfig:adj_acc_workers}}} = 
    \sum_{k=0}^{n} 
    \underbrace{\evb{\f(x_2)\vert \hs{x_2} = \frac{k}{n}}}_{
    \text{Fig. \ref{subfig:p_f_given_s}}} \cdot 
    \underbrace{p_1\left(\hs{x_1} =\frac{k}{n}\right)}_{
    \text{Fig. \ref{subfig:ps_workers_raw}}},
\end{align}
and allows us to draw the following conclusions:
\begin{itemize}
    \item Figure~\ref{subfig:ps_workers_raw} shows that the distribution
    of observed \va selection frequencies $p_1(\hs{x})$ becomes increasingly skewed as 
    more annotators are used to estimate selection frequencies (i.e. as bias decreases). 
    \item Figure~\ref{subfig:p_f_given_s} plots selection frequency-conditinoed
    classifier accuracy,  $\evb{\f(x_2)\vert \hs{x_2} = \frac{k}{n}}$ as a
    function of $n$. The plot indicates
    that when we use observed selection frequency in place of true selection
    frequency, we overestimate model accuracy on images with low selection
    frequency and underestimate accuracy on images with high selection frequency.
    \item Combining these two sources of bias,
    Figure~\ref{subfig:adj_acc_workers} shows that as we reduce bias by 
    increasing $n$, the selection-adjusted \vb{} accuracy increases for every
    classifier.  
\end{itemize}

It turns out that computing~\eqref{eq:adjacc_hat} using the 40 annotators per
image that we collected in Section~\ref{sec:remeasure} already produces
selection-adjusted \vb~accuracies that are on average 6.0\% higher than
the initially observed \vb accuracy. Thus, despite still suffering from
statistic matching bias, the na\"ive approach reduces the \va-\vb{} accuracy drop to
5.7\% using the remeasured selection frequencies. 
In the following sections, we explore two different techniques for debiasing the
na\"ive estimator and explaining more of the accuracy gap.

\subsection{Estimating bias with the statistical jackknife}
\label{sec:jackknife}
As a first attempt at correcting for the previously identified bias, we turn to a
standard tool from classical statistics. The
jackknife~\citep{quenouille1949plane,quenouille1956notes,tukey1958bias} is
a nonparametric method for reducing the bias of finite-sample estimators. 
Under mild assumptions, the jackknife 
can reduce the bias of any estimator from $O(\frac{1}{n})$ to $O(\frac{1}{n^2})$. 
Concretely, the jackknife bias estimate for an $n$-sample estimator $\theta_n :=
\hat{\theta}(X_1,\ldots,X_n)$ is given by:
\begin{align*}
b_{jack}(\hat{\theta}_n) &= (n-1)\cdot \left(\frac{1}{n}\sum_{i=1}
\hat{\theta}_{n-1}^{(i)} - \hat{\theta}_n\right), \\
\text{where}\qquad \hat{\theta}_{n-1}^{(i)} &=
\hat{\theta}(X_1,\ldots,X_{i-1},X_{i+1},\ldots,X_n) \text{ is the $i$th
leave-one-out estimate.}
\end{align*}
The statistical jackknife thus provides us with a technique for
estimating and correcting for the bias in finite-sample estimates of the
frequency-adjusted accuracy $\adjacc$.  

\paragraph{Jackknifing the na\"ive estimator. } As a first approach, we can
apply the jackknife directly to the na\"ive estimator
(cf.~\eqref{eq:adjacc_hat}). For the jackknife-corrected estimate to be
meaningful, we have to show that the na\"ive estimator is a statistically
consistent estimator of the true selection-adjusted accuracy (i.e., that
$\lim_{n\rightarrow\infty} \naiveest = \adjacc$). We prove this property in
Appendix~\ref{app:kde}, under the assumption that we can evaluate quantities of
the form $p_i(\hs{x} = s)$ exactly (in practice this assumption seems acceptable since the
empirical variance of the estimator is small)\footnote{It is also possible to
prove consistency even without this assumption by assuming a fixed
relationship between the number of images and the number of annotators per
image, and at the cost of the proof's simplicity.}. Applying the jackknife to
the na\"ive estimator reduces the adjusted accuracy gap further, from {\bf
5.7\%} to {\bf 4.6\%}.

\begin{figure}
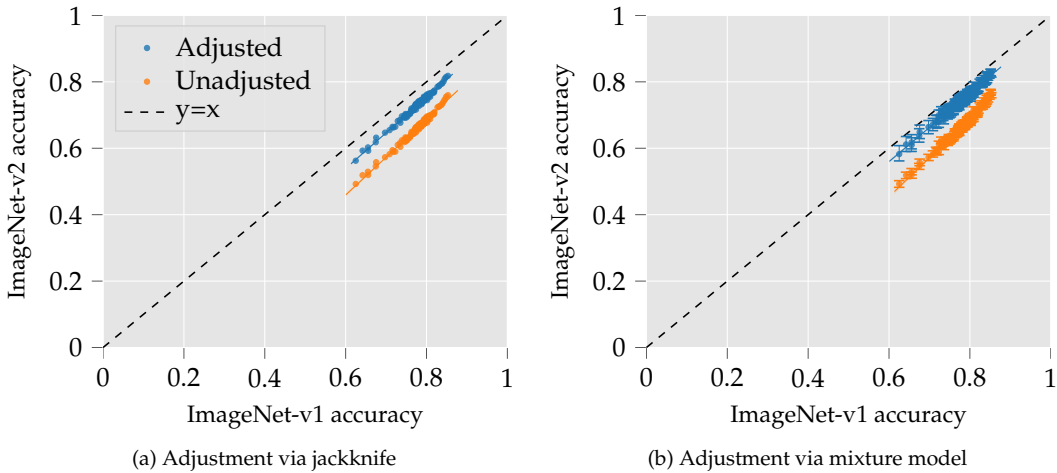

    \centering
    \subfloat[Adjustment via jackknife]{
        \input{plots/jackknife_clean}
        \label{fig:jackknife_results}
    }
    \subfloat[Adjustment via mixture model]{
        \input{plots/boostrap_betabinom_zoomout}
        \label{fig:bb_results}
    }
    \caption{Accuracy on \va~versus \vb adjusted using the two techniques
        discussed in this section. On the left (respectively, right) we use
        the jackknife (parametric model) of
        Section~\ref{sec:jackknife}~(\ref{sec:parametric}) to estimate
        adjusted accuracies for \vb. The graphs confirm that the ``true'' gap
    in accuracy between \va~and \vb~is indeed much smaller than the
    initially observed gap. Confidence intervals on the left are based on the
jackknife standard error, and confidence intervals on the right are based on 400-sample 95\%
bootstrap confidence intervals}
\end{figure}

\paragraph{Considerations and limitations of the jackknife approach. } For
the jackknife to perform reliably, we must have that (a) the leave-one-out
estimators have low enough variance, and (b) the bias is an analytic function in
$1/n$ that is dominated by the $\Theta(\frac{1}{n})$ term in its power series
expansion. We address the first of these concerns by plotting
jackknife confidence intervals (c.f.~\cite{efron1994introduction}) for our
estimates. Consideration (b) carries a bit more weight: as shown in
Appendix~\ref{app:jk_justification}, the $n$-sample na\"ive estimator has a
roughly linear relationship in $1/n$, but not a perfect one---in particular, the
estimator seems to increase at a rate slightly faster than $1/n$,
suggesting that as a result, the jackknife still provides an underestimate of
the selection-adjusted accuracy.

Another potential source of error is finite-sample error in measuring the
expectations $\evb{\f(x_2)\vert 
\hs{x_2}}$. If the number of workers per image $n$ is
taken to infinity while keeping the number of images constant, the observed
selection frequencies will become too sparse to provide reliable estimates of the
expectation. Thus we rely on our assumption from
Section~\ref{sec:exps} that finite-sample error is negligible with respect to
images (since we have $10^4$ images and only $41$ possible values of
$\hat{s}$). 
Also, note that in principle, we
could also use the jackknife to estimate the distributions of true selection
frequencies ($p_i(\s{x})$) by estimating the bias in the statistic
$\mathbb{E}[\mathbbm{1}_{\hs{x} = k}]$ for each value of $k$. However, this
approach is too sample-expensive to yield reliable results.

In the next section, we present another approach to estimating the
selection-adjusted accuracy, namely parametric modeling. Through a disjoint set
of techniques and assumptions, we obtain estimates of $p_i(\s{x})$ and $\adjacc$
that further corroborate the results so far.

\subsection{Estimating bias with a parametric model}
\label{sec:parametric}
We now explore a more fine-grained approach to estimating the selection-adjusted
accuracy of ImageNet-v2, namely 
explicit parametric modeling. Recall that the adjusted accuracy captures
accuracy on ImageNet-v2 reweighted to match ImageNet-v1 in terms of true
selection frequency distribution, and is given by:
\begin{align}
    \nonumber
    \adjacc &= \int_s \mathbb{E}_{\dvb}[f(x)|s(x)=s] \cdot p_1(s)\ ds \\
    &= \int_{s \in [0,1]} p_2\left(f(x_2)\vert\s{x_2} = s \right)
    \cdot p_1(s)\ ds 
    \label{eq:prods}
\end{align}
In Sections~\ref{sec:naive} we computed a biased estimate of $\adjacc$
using observed selection frequencies $\hat{s}$ in place of true selection
frequencies. Then, in~\ref{sec:jackknife} we corrected for the bias in this
na\"ive estimator post-hoc using the statistical jackknife. 

In constrast, the model-based approach tries to circumvent this bias altogether:
we parameterize functions of the true selection frequency directly (i.e.,
$p_1(s)$ and $p_2(f(x) = 1 \vert s(x) = s)$), then fit parameters that maximize
the likelihood of the observed data while taking into account the noise model.  
For example, since the distribution of $\hs{x}$ given $\s{x]}$ is the
binomial distribution, we can write (and optimize) a closed-form expression for
the likelihood of observing a given set of selection frequencies based on 
a parameterized true selection frequency distribution $p_1(s;\theta)$.
We estimate selection-adjusted accuracy in two steps. First, we fit models for
the true selection frequency distributions $p_1(s)$ and $p_2(s)$. Then, we
use our estimate of $p_2(s)$ in conjunction with observed data to fit
models for $p_2(f(x) \vert \s{x} = s)$. Finally, we recover estimates for
$\adjacc$ by numerically computing the integral in $\adjacc$
(c.f.~\eqref{eq:prods}), plugging in the learned parametric estimates.

\paragraph{Fitting a model to $p_i(\s{x})$.}
We model the $p_i(\s{x})$ as members of a parameterized family of
distribution $p_i(\s{x}; \theta)$ with true parameters $\theta_i^\star$.
Then, for each dataset $i$, we model the observed selection frequencies as
sampled from a compound distribution, in which one first samples $s \sim
p_i(\cdot;\theta_i^\star)$, then observes $\hat{s} \sim \binomial{s}$ (where
$n$ is the number of MTurk annotators).

To infer each $\theta_i^\star$, we use maximum likelihood estimation on the
observed samples over the compound distribution. We opt to use
mixtures of beta distributions as the family $p_i(\cdot;\theta)$ over
which to optimize. Beta-mixture distributions are a fairly popular modeling
choice for finite-support
data~\citep{ji2005applications,ma2009beta,laurila2011beta}, since beta
distributions (a) have only two parameters; (b) induce smooth, continuous
density functions with support $[0, 1]$; and (c) admit a closed-form likelihood
when composed with a binomial random variable, by way of the {\em beta-binomial}
distribution\footnote{\url{https://docs.scipy.org/doc/scipy/reference/generated/scipy.stats.betabinom.html}}.

\begin{figure}
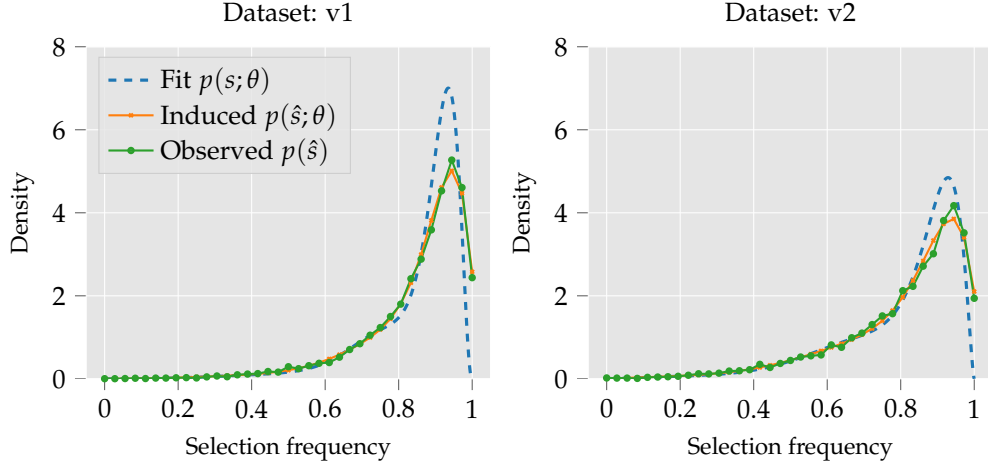

    \centering
    \input{plots/bb_ps_clean_v1.tex} 
    \input{plots/bb_ps_clean_v2.tex} 
    \caption{Our
    fit beta mixture models $p_i(s; \hat{\theta})$ for ``true'' selection
    frequency, the noisy selection frequency distribution they induce
    $p_i(\hat{s}; \hat{\theta}$, and the observed selection frequency
    $p_i(\hat{s})$. We observe that the fit $p_i(s; \hat{\theta})$ distributions
    place much more density on higher selection frequencies than na\"ively
    estimating $p_i(s)$ from the observed $p_i(\hat{s})$.} \label{fig:fit_betas}
\end{figure}

A mixture of $k$ beta distributions is a $3k-1$-parameter model; we denote
the mixture coefficients by $\gamma_j$ (with $\sum_j \gamma_j = 1$), and
the parameters of each individual beta distribution in the mixture as
$(\alpha_j, \beta_j)$. We use Expectation-Maximization (EM)~\citep{dempster1977maximum}
to find, for each dataset $i \in \{1, 2\}$, the maximum likelihood mixture
of $k=3$ beta-binomial distributions for the observed $\hs{x}$. The
log-likelihood is given by:
\begin{equation*}
    \hat{\theta}_i := \bm{\beta}^{(i)}, \bm{\alpha}^{(i)}, \bm{\gamma}^{(i)} = 
\argmax_{\bm{\beta}, \bm{\alpha} \in \mathbb{R}^{3}_{+}, 
    \bm{\gamma} \in \Delta^{3}}\ \ \ 
    \sum_{x \sim \dvi} \log\left( \sum_{j=1}^3 \gamma_j \bbn\left(\alpha_j, \beta_j, N, x\right)\right),
\end{equation*}
where $\bbn(\alpha, \beta, N, x)$ is the density of the beta-binomial
distribution parameterized by $(\alpha, \beta, N)$ and evaluated at $x$. We
provide
further detail on the fitting process for $p_i(\s{x};\theta)$ in
Appendix~\ref{app:model_fitting}, including pseudocode for the EM
algorithm. We plot the resulting fitted distributions $p_i(\s{x})$ in
Figure~\ref{fig:fit_betas}. Our estimated $p_i(\hat{s}; \hat{\theta}_i)$
distributions continue the trend previously seen in
Figure~\ref{subfig:ps_workers_raw}, and show the extent to which our na\"ive
40-sample empirical estimates of $p_i(\s{x})$ exhibit bias.

\paragraph{Fitting a model to $p_2(f(x) = 1 \vert \s{x} = s)$.}
Next, we consider accuracy conditioned on selection frequency:
$$g(s) = p_2(f(x_2) = 1 \vert s(x_2) = s).$$ 
While introducing the na\"ive estimator (Section~\ref{sec:naive}), we found that
that estimating $g(s)$ using observed selection frequencies instead of true
selection frequencies results in bias (Figure~\ref{subfig:p_f_given_s}). Under
the parametric approach, we instead 
model $g(s)$ as a member of a parametric class (i.e., $g(s) = g(s; \omega)$),
then account for noise in observed selection frequencies via the following
identity:
\begin{equation}
    p_2(\f(x_2)=1, \hs{x_2}=\hat{s}) = \int_{s \in [0, 1]} g(s)\cdot p(\hat{s}\vert s)\cdot p_2(s)\ ds. 
    \label{eq:joint_ident}
\end{equation}
Now, since $g$ is an unknown but most likely smooth
function from $[0,1]$ to $[0,1]$, a standard parameterized family
$g(\cdot;\omega)$ to use is the class of cubic splines~\citep{de1978practical}.
In particular, we try to find a $g(s;\omega)$ satisfying the relationship
described by \eqref{eq:joint_ident} by optimizing the following squared error
objective:
\begin{equation}
    L(\omega) := \sum_{\hat{s}} \left( p_2(\f(x_2)=1, \hs{x_2} = \hat{s})
    - \int_{s \in [0, 1]} g(s;\omega)\cdot p(\hat{s} \vert s)\cdot
    p_2(s;\hat{\theta}_2) ds \right)^2.\label{eq:obj}
\end{equation}
We can compute the left hand term above from observed data, and the right-hand
term is a function of the parametrized $g(s;\omega)$, the binomial mass
function, and $p_2(s;\hat{\theta}_2)$, which we estimated in the first step.
After discretizing the integral,~\eqref{eq:obj} becomes a quadratic optimization
problem \footnote{Note the choice of least-squares loss here is entirely for
convenience---in principle one could maximize the likelihood of
$p_2(\f(x_2)=1,\hs{x_2}=\hat{s})$ instead at the cost of algorithmic simplicity
and efficiency} (since splines are linear in their parameters).

\paragraph{Results.} Once we have estimated probability
distributions $p_i(\s{x};\theta_i)$ and the conditional classification function
$g(\s{x}; \omega)$, we can compute an estimate of $\adjacc$ using
Equation~\eqref{eq:prods} and numerical integration. Figure~\ref{fig:bb_results}
depicts various models' \va~and \vb~accuracies both with and without the
adjustment for (remeasured) selection frequency. Our estimate for the frequency-adjusted gap
in accuracy averaged over all models is $3.6\% \pm 1.5\%$, around 30\% of the
original $11.7\% \pm 1.0\%$ gap in accuracy. 

Beyond accuracy gap, \citet{recht2018imagenet} also studied the linear
relationship between \vb accuracy and \va accuracy while varying the classifier
used---this is plotted by the blue dots in Figure~\ref{fig:bb_results}. This
relationship is linear for our adjusted accuracies as well (cf.
Figure~\ref{fig:bb_results}), however the slope we find is $1.01 \pm 0.09$ 
instead of $1.13 \pm 0.05$.

\paragraph{Considerations and limitations.}
Error in parametric modeling generally stems from two sources: finite-sample
error and model misspecification. These sources of error affect
all parametric models, but we take various precautions to mitigate their impact
on our estimates. 

To assess our finite-sample error, we give 95\% bootstrapped confidence intervals
(details are in Appendix~\ref{app:model_fitting}), which are displayed as error
bars in Figure~\ref{fig:bb_results}. We also
ensure that our results are not sensitive to the number of annotators used to
fit the parametric models (cf. Appendix~\ref{app:model_fitting}).

As with any modeling decision, our choices of model classes might not fully
capture the ground-truth generative process, and thus may be a source of
error. We account for this as much as possible by demonstrating the
robustness of our results to varying the number of free parameters (cf.
Appendix~\ref{app:model_fitting}).

\section{Related Work}
\label{sec:rel_work}

Researchers have developed a cornucopia of datasets (e.g.~\citep{lecun1998mnist,
  krizhevsky2009learning, russakovsky2015imagenet, zhou2017places} in computer
vision) for benchmarking the generalization performance of supervised learning
algorithms. The recent rapid pace of progress has drawn interest to the
question of
verifying that the progress made by high-performing models on standard dataset
actually corresponds to progress on the underlying task, rather than on the
dataset itself. To this end, previous work has characterized cross-dataset
generalization on similar tasks~\citep{torralba2011unbiased} by measuring
``dataset bias.'' Prior work has also explored the impact of synthetic
perturbations on generalization, such as adversarial
examples~\citep{kurakin2016adversarial,tsipras2019robustness,ilyas2019adversarial,su2018robustness}
or various other corruption robustness
measures~\citep{hendrycks2019benchmarking,kang2019testing}.

In terms of measuring \textit{same}-distribution generalization beyond test set
performance, a number of works have emerged around evaluating performance on
newly reproduced test sets, including works focusing on ImageNet~\citep{recht2018imagenet},
CIFAR~\citep{recht2018imagenet}, and MNIST~\citep{yadav2019cold}. By and large,
these works report accuracy drops due to distribution shift, and claim that
there is no adaptive overfitting.

Adaptively reusing data is a studied source of both overfitting in machine
learning and, more generally, false discovery in the
sciences~\citep{gelman2014statistical}. In machine learning, work has gone into
theoretically characterizing and counteracting the effects of adaptive
overfitting~\citep{blum2015ladder, dwork2015generalization, roelofs2019meta}. In
computer vision in particular, \citet{recht2018imagenet} and
\citet{mania2019model} respectively give evidence that there is no adaptive
overfitting, and that model similarity reduces adaptive overfitting.

A common phenomena noted in ecology is
overdispersion~\citep{greig1983quantitative}, a form of sampling bias that can
emerge when one does not accurately model the underlying distribution generating
samples. The bias we discuss can be framed as an instantiation of this problem.

\section{Discussion and Conclusions}
\label{sec:implications}
Dataset replication pipelines can introduce unforeseen, often unintuitive
statistical biases. In the case of ImageNet-v2, even using unbiased estimates of
image selection frequency in the data generation pipeline results in a
significant statistical bias, and ultimately turns out to account for a large
portion of the observed accuracy drop. Our findings give rise to the following
considerations.

\subsection{Remaining accuracy gap and unmodeled bias}
\paragraph{Worker heterogeniety.} Our study focuses on bias stemming from the
fact that for a given image $x$ one never observes $\s{x}$ but rather $\hs{x} =
\binomial{\s{x}}$. There is another source of bias due to noise that we
do not model here, namely variance in the MTurk annotator population.
Specifically, some annotators are more likely in general to select or reject
independently of what image-label pair they are being shown. This unmodeled
variance likely translates to unmodeled bias, suggesting that more of the gap
might be explained by taking worker heterogeniety into account.

\paragraph{Task shift bias.} At the time of the original ImageNet experiment,
workers judged image-label pairs by some abstract set of criteria $C_1$. Suppose
that at the time of the ImageNet-v2 experiment several years later, annotators
judged image-label pairs based on an overlapping but non-identical set of
criteria $C_2$. Ideally, we should not care about differences between $C_1$ and
$C_2$---indeed, one of the goals of dataset replication is to test robustness to
such benign distribution shifts. The source of the bias lies in the iterated
nature of the filtering experiment. In particular, after both the original
experiment and the replication, images in ImageNet-v1 now meet both $C_1$ and
$C_2$. On the other hand, images in ImageNet-v2 only meet criteria $C_2$, and
may be judged to have low selection frequency under $C_1$---we would thus expect
models to perform better on ImageNet-v1 images due to their increased
qualifications. Although this may contribute towards the remaining accuracy gap,
this type of bias is difficult to study or correct for without more knowledge of
both experiments.

\paragraph{Other sources of error.} The remaining error unexplained by 
bias in data collection could come from one of the gap sources listed in
Section~\ref{sec:exps}, i.e., finite sample error, or distribution shift
and adaptive overfitting. 
Quantifying the potential contribution of the individual terms in the remaining gap
will require more experimentation and future work.

\subsection{Adaptive overfitting and distribution shift}
\paragraph{Identifying sources of distribution shift.} A longstanding goal in
computer vision is to develop models that are less prone to failure under small
distributional shifts. A step in the journey towards this goal is precisely
characterizing the kinds of distribution shifts under which models
fail---examples include rotations and translations of natural
images~\citet{engstrom2019rotation}, or corrupted natural
images~\citep{hendrycks2019benchmarking}. Our findings imply that the drop may
be attributable to differences in selection frequency distribution,
corroborating observations by~\citet{recht2018imagenet} that models are
sensitive to selection frequency. Differences in selection frequency
distribution present another distribution shift to study in depth.

\paragraph{Implications for adaptive overfitting.} A conclusion drawn
from the ImageNet-v2 dataset replication is that 
the slope of best fit line for \va{} versus \vb{}
accuracy is significantly larger than one. This means that for every point of
progress made on ImageNet, $1.1$ points are made on ImageNet-v2, providing
another point of evidence towards the absence of adaptive overfitting. However,
after correcting for the bias in the \vb{} sampling process, \va{} versus \vb{}
model accuracies still exhibit a linear fit, but with a slope that is not
conclusively bounded away from $1$ (i.e., slopes are within 95\% confidence
intervals of $1$, c.f. Section~\ref{sec:parametric}). 

\paragraph{Detecting and avoiding bias in dataset replication.} 
More broadly, our analysis identifies statistical modeling of the data
collection pipeline as a useful tool for dataset replication. Indeed,
characterizing the ImageNet and ImageNet-v2 generative processes and isolating
them in a a simple theoretical model allowed for the discovery and correction of
a source of bias in the dataset replication process.

\section*{Acknowledgements}
We thank Will Fithian for discussions and advice, particularly around the
spline modeling done in Section~\ref{sec:quantifying}. We also thank Nicholas
Carlini, Zachary Lipton, Kevin Foley, and Michael Yang for helpful comments on
early drafts of this paper. Finally, we thank Nur Muhammad (Mahi)
Shafiullah for useful discussions and suggestions early in our
investigation.

Work supported in part by the NSF grants CCF-1553428, CNS-1815221, the Google
PhD Fellowship, the Open Phil AI Fellowship, and the Microsoft Corporation.

Research was sponsored by the United States Air Force Research Laboratory and
was accomplished under Cooperative Agreement Number FA8750-19-2-1000. The views
and conclusions contained in this document are those of the authors and should
not be interpreted as representing the official policies, either expressed or
implied, of the United States Air Force or the U.S. Government. The U.S.
Government is authorized to reproduce and distribute reprints for Government
purposes notwithstanding any copyright notation herein.

\pagebreak
\printbibliography

\pagebreak
\appendix
\section{Understanding Selection Frequency}
\label{app:understanding_sel_freq}
\begin{figure}[!ht]
\centering
\includegraphics[width=0.75\textwidth]{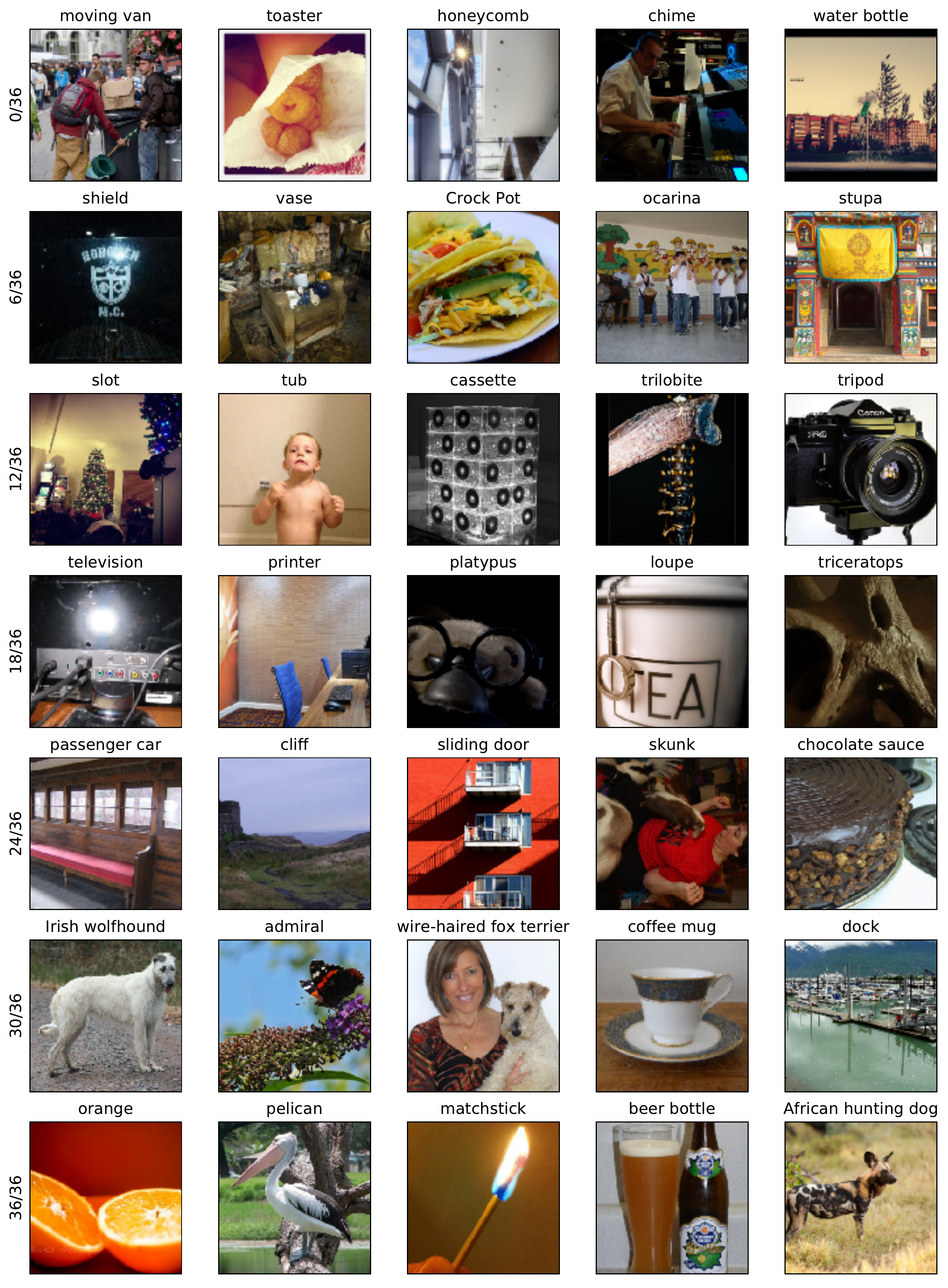}
\caption{Randomly drawn images from \va, varying selection frequency.}
\label{app:sel_freqs}
\end{figure}

\paragraph{Understanding Selection Frequencies}
In Figure~\ref{app:sel_freqs} we randomly sample images while varying selection
frequency. Here, the straightforwardness of identifying images correlates with increasing selection frequency (e.g. all the 36/36 selection frequency images clearly identify with their corresponding class, while some of the 0/36 selection frequency images appear to be mislabeled).

\section{Experimental Setup}
\label{app:exp_setup}
Here, we provide more detail of the experimental setup. We first lay out the
setup of our Mechanical Turk experiments for remeasuring selection frequency
(\ref{app:remeasure_setup}), and highlight the subtle differences between our
setup and that of~\citet{recht2018imagenet} (\ref{app:exp_differences}). In
Appendix~\ref{app:orig_data} we discuss our analysis of the original data and
algorithm of~\citet{recht2018imagenet} showing the existence of bias in that
setting. 

\subsection{Selection frequency remeasuring experiment}
\label{app:remeasure_setup}
In Section~\ref{sec:remeasure}, we replicate the ImageNet experiment to
remeasure selection frequencies for the ImageNet v1 and v2 datasets. We present
annotators with a grids of 48 images along with an ImageNet class. The
annotators are also provided with the WordNet synsets for the ImageNet class
being queried, along with a Wikipedia link and asked to selected all images
containing instances of that object (ignoring clutter as per the original
dataset creation process). The 48 images in each grid consist of: (a) 10
ImageNet v1 validation set images from that class, (b) 10 ImageNet v2 validation
set images from that class, (c) 22 related Flickr images scraped from Flickr
(using the exact script and queries described in \citet{recht2018imagenet}) and
(c) 6 negative control images (three corresponding to randomly chosen labels,
and 3 corresponding to the ``nearest'' label to the true label in terms of
WordNet path similarity).

\begin{figure}[!ht]
	\centering
	\includegraphics[width=0.85\textwidth]{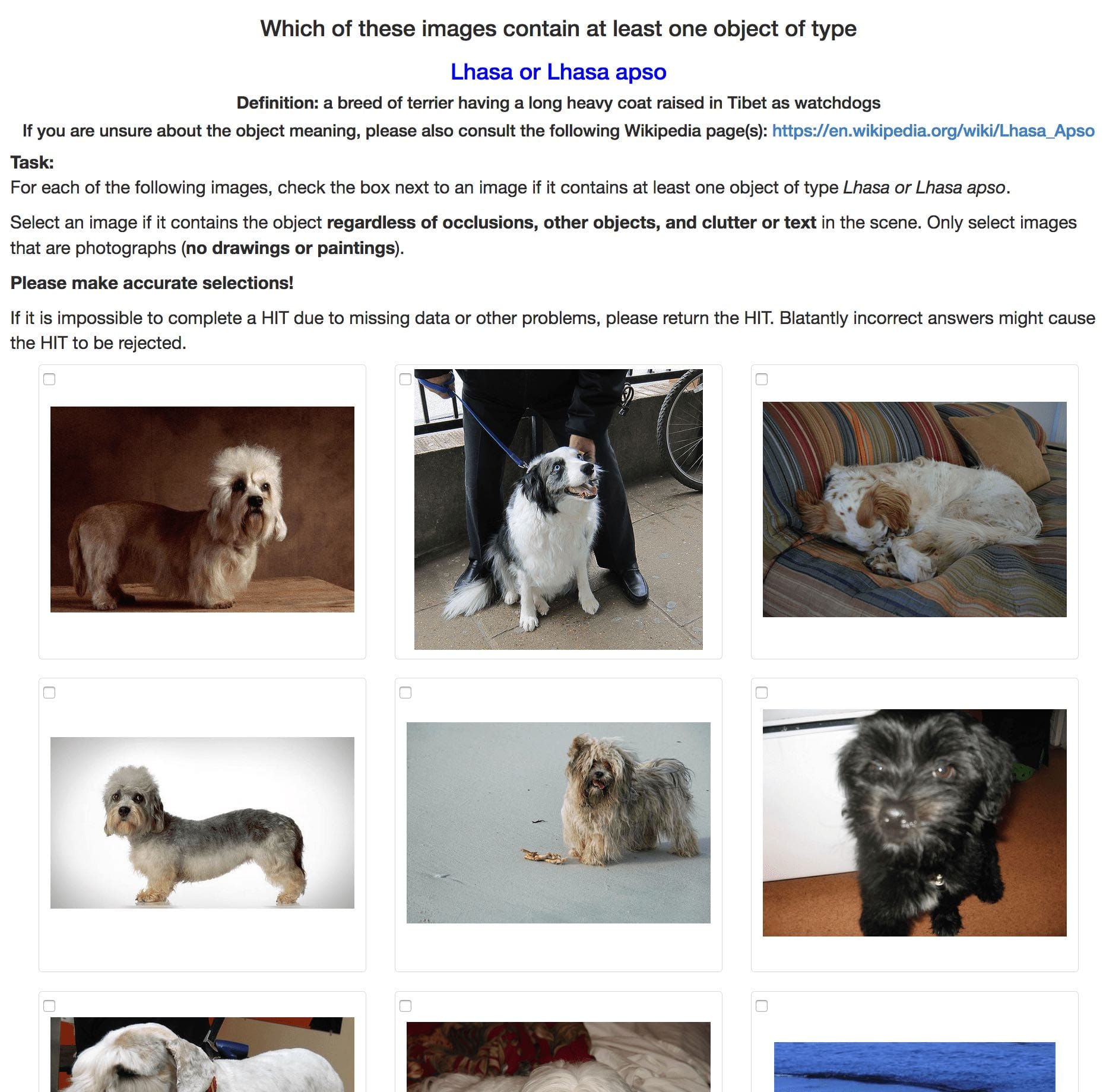}
	\caption{Screenshot of a sample grid to measure selection frequencies
		(Section~\ref{sec:remeasure}).
		Annotators are given a grid of 48 images for a specific label and asked to
		select all images that contain that label. The interface and instructions
		are based on
		\citet{recht2018imagenet}.}
	\label{fig:contains_screen}
\end{figure}

We implemented our setup by modifying the code made publicly available by
\citet{recht2018imagenet}~\footnote{\url{https://github.com/modestyachts/ImageNetV2}}.
A screenshot of our interface appears in Figure~\ref{fig:contains_screen}. Each
such grid of images is shown to 40 annotators. For each image-class pair, we can then compute the ``selection frequency''
based on how often it was selected by the annotators.

\paragraph{Deployment Details.}
There are a number of deployment details that could cause variations in results.
We compensated MTurk workers with \$0.23 per assignment (i.e., each completed
grid), which we calibrated to pay a rate of at least \$9/hr for most workers. To
collect 40 separate MTurk annotations for each submitted grid of images, we
obtained 10 annotations on 4 different dates and times, all within the span of
a single week. We placed qualification requirements on the workers allowed to
complete assignments. Specifically, we filtered for workers that (a) agree to
view adult content (as some ImageNet images have content like nudity or gore)
and (b) have a larger than 95\% assignment approval rate (as to ensure the
quality of the results).

\paragraph{Controls.} All of the results presented in this work were run on a
``clean'' and ``raw'' version of our data, i.e., without and with data cleaning
respectively. We find that the inclusion of data cleaning makes the observed gap
between \va and \vb slightly larger but otherwise does not have a significant
effect on results.

Our data cleaning process is as follows: a given batch is
``flagged'' if: (a) there are less than 6 selected images out of the total 48,
or (b) more than one of the negative controls was selected. We only omit
data, however, from workers whose batches were rejected at a rate of 30\% or
higher (e.g., if an annotator completed 30 batches, but more than 10 of them are
flagged to be low-quality, then all of the annotator's data is omitted).
Finally, to make computing of the statistics easier, we evened out the number
of annotators per image to equal the minimum number of remaining assignments
per image, which was 36 (compared to 40 originally) by randomly discarding
annotations. In total, the entire process corresponds to discarding 10\% of the
annotations.

\subsection{Comparison to the original setup}
\label{app:exp_differences}
\citet{recht2018imagenet} measure the average selection frequency of \va to be
0.71, whereas our experiment measures the average \va selection
frequency to be 0.85. While our experiments were modeled closely after that
of~\citet{recht2018imagenet} (and in fact use the same core codebase to minimize
discrepancies in task presentation/inferace), we made a few changes to the setup
to ensure high data quality. We hypothesize that these changes, discussed below,
are what result in the discrepancy between the measured average selection frequencies.
However, since these changes are applied at the task level and annotators are
not told which dataset each image is sourced from, we find it unlikely that these changes
would affect annotations for one dataset more than the other. 
Furthermore, in Appendix~\ref{app:orig_data}, we demonstrate that the bias
identified in this paper can be found even using the original data collected
by~\citet{recht2018imagenet}.

\paragraph{Worker pay and qualifications.} In our experiment, we
paid annotators 20 cents per set of 48 images completed---this was informed by
the average time taken to complete a batch, and was calibrated so that the task
paid approximately 12 dollars per hour. Conversely, the original experiment
of~\citet{recht2018imagenet} pay 10 cents per batch. Although worker pay usually has
only a mild impact on worker reliability on
MTurk~\citep{mason2009financial,buhrmester2011amazon}, higher worker pay has
been recognized as a tool to boost participation rates~\citep{buhrmester2011amazon} and requester
reputation for future experiments~\citep{paolacci2010running}.

Perhaps the most important modification made was our inclusion of worker
qualifications, which only allow annotators who have had 95\% or more of
previous tasks accepted to participate in our task. Prior work has
shown that without these worker qualifications,
crowdsourced data tends to be of significantly worse quality. For
example,~\citet{peer2013reputation} report that 2.6\% of workers with the ``95\%
accepted'' qualification failed an ``attention-check
test,'' compared to 33.9\% of workers without
qualifications\footnote{Attention-check tests are a series of three
attention-check questions (ACQs). ACQs are
questions with right/wrong answers unrelated to the task meant to gauge an
annotator's attentiveness, e.g. ``Have you ever had a fatal heart attack?''. In
the~\citet{peer2013reputation} study, 16.4\% of unqualified workers reported
that they had suffered a {\em fatal} heart attack, compared to 0.4\% of
qualified workers.}. We should therefore expect a significant increase in
annotation reliability (and so in turn some discrepancy) from using
worker qualifications.

\paragraph{Makeup of each batch.} Another difference between the two experiments
is that in our experiment, each batch of images contains 10 images from
ImageNet-v1, and 10 images from ImageNet-v2, in order to ensure that we could
obtain 40 annotations for each \va and \vb image while keeping to a reasonable
budget constraint. The experiment of~\citet{recht2018imagenet} uses only
three images from ImageNet-v1 per batch (and a variable number of ImageNet-v2
images, since the dataset was not yet realized). Thus, the grids presented in our
experiment contain images that are on average more likely ({\em a priori}) to be
selected. This could in part contribute to the higher average selection
frequency that we observe (though again, we would expect this effect to apply
to both datasets and thus preserve the observed selection frequency gap).

\paragraph{Randomization.} Response-order bias is a well-documented
phenomenon in literature on crowdsourcing (e.g.~\citep{schuman1981question}),
although its effects in the domain of image selection are not
well-understood yet. In our experiment, we randomize the order of the images per
batch per worker (i.e., we used JavaScript to randomize the image order on
page load) to mitigate the potential effects of this bias. In the prior
experiment, however, the image order is deterministic, and thus the study may
have response-order effects.

\begin{figure}[hb]
    \centering
    \begin{tikzpicture}

\definecolor{color0}{rgb}{0.12156862745098,0.466666666666667,0.705882352941177}

\begin{axis}[
height={4cm},
width={0.6\textwidth},
axis background/.style={fill=white!89.8039215686275!black},
axis line style={white},
log basis y={2},
tick align=outside,
tick pos=left,
x grid style={white},
xlabel={\small Number of copies of the same (worker, batch) pair},
xmajorgrids,
xmin=0.31, xmax=6.69,
xtick style={color=white!33.3333333333333!black},
y grid style={white},
ylabel={\small Count},
ymajorgrids,
ymin=1, ymax=2184025.29151296,
ymode=log,
ytick style={color=white!33.3333333333333!black}
]
\draw[draw=none,fill=color0,very thin] (axis cs:0.6,0) rectangle (axis cs:1.4,1148440);
\draw[draw=none,fill=color0,very thin] (axis cs:1.6,0) rectangle (axis cs:2.4,28942);
\draw[draw=none,fill=color0,very thin] (axis cs:2.6,0) rectangle (axis cs:3.4,1287);
\draw[draw=none,fill=color0,very thin] (axis cs:3.6,0) rectangle (axis cs:4.4,208);
\draw[draw=none,fill=color0,very thin] (axis cs:4.6,0) rectangle (axis cs:5.4,25);
\draw[draw=none,fill=color0,very thin] (axis cs:5.6,0) rectangle (axis cs:6.4,3);
\end{axis}

\end{tikzpicture}
    \caption{A histogram showing the existence of duplicate (worker, batch)
    pairs present in the original collected data. Each point in the histogram
    is a unique (worker, batch) pair, and the $x$ axis corresponds to the
    number of times that pair is observed in the dataset.}
    \label{fig:worker_image_hist}
\end{figure}
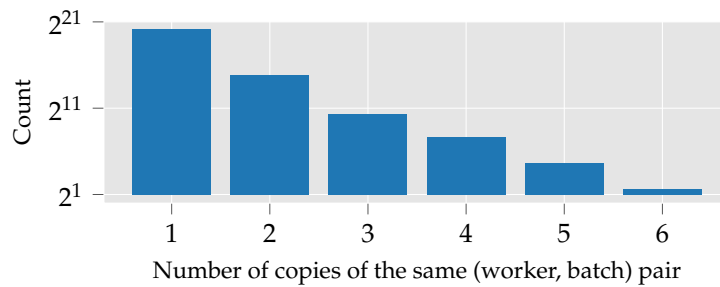

\paragraph{Worker duplicates.} 
Due to the mechanism by which images were distributed to assignments in the study
of~\citet{recht2018imagenet}, 5.6\% of the annotations are duplicated (i.e.,
5.6\% of the [worker, image] pairs collected are non-unique), with approximately
3\% of the annotations being redundant (unlike the preceding number, this fraction does
not count the ``original copy'' of each non-unique pair). A histogram of the
number of times a single worker labeled a single image is shown in
Figure~\ref{fig:worker_image_hist}. Since duplicate workers violate
sample indpendence and can skew measured selection frequencies for some images,
in our study we ensure that no worker labels the same image more than once.

\paragraph{Data cleaning and controls.} Our study also differs in having
built-in mechanisms for data cleaning (as discussed in the last section),
allowing us to run all of our experiment on the ``cleaned'' and ``raw'' versions
of our data. These results tend to not be substantially different (for the
cleaned data, the selection frequency gap we measure between \va and \vb 
slightly increases from $4.5\%$ to $4.6\%$). Possible reasons for this
similarity between cleaned and raw results include any of the quality control
protocols outlined in this section.

\clearpage

\section{Theoretical Results}
\label{app:theory_model}
In this section we show the series of calculations used in
Section~\ref{sec:id_bias} to attain the result in
Equation~\ref{eq:toymodelsoln}, i.e. the bias incurred by a matching
procedure in the toy model.

Recall that in our setup we have $p_{flickr}(s)$ and $p_{1}(s)$ given by
$\text{Beta}(\alpha,\beta)$ and $\text{Beta}(\alpha+1,\beta)$
respectively, and that $\hat{s}(x)$ is given by first sampling $s \sim
p_i(s)$ then sampling $n$ Bernoulli trials with success probability $s$. We
noted in Section~\ref{sec:id_bias} that the distribution of $s(x)$ induced
by matching $p_{flickr}(s)$ and $p_{1}(s)$ based on samples of $\hat{s}(x)$
is given by:
\begin{align*}
p_{flickr}(s(x)) \cdot \mathbb{P}(\text{x is accepted}\vert s(x)) &=
p_{flickr}(s(x)) \cdot \int_{\hat{s}} p(\hat{s}\vert s) \mathbb{P}(\text{x
is accepted} \vert \hat{s}(x)) \\
&= p_{flickr}(s(x)) \cdot \int_{\hat{s}} p(\hat{s}\vert s)
\frac{p_{1}(\hat{s}(x))}{p_{flickr}(\hat{s}(x))}
\end{align*}

Now, note that by construction, $\hat{s}(x)$ is distributed according to
the beta-binomial
distribution\footnote{\url{https://en.wikipedia.org/wiki/Beta-binomial_distribution}},
and thus (a) has support $\{0,\ldots,n\}$; and (b) induces the following
closed-form probability mass function for $p_{flickr}(\hat{s})$
($p_{1}(\hat{s})$ can be written analogously, with $\alpha+1$
replacing $\alpha$) :
$$p_{flickr}(\hat{s}(x) = k) = {n \choose k} \frac{B(k+\alpha, n-k+\beta)}{B(\alpha,
\beta)},$$
where 
$$B(\alpha, \beta) =
\frac{\Gamma(\alpha)\Gamma(\beta)}{\Gamma(\alpha+\beta)},$$
and $\Gamma$ is the Gamma function---for simplicity we will assume that
$\alpha, \beta \in \mathbb{N}$ and so $\Gamma(x) = (x-1)!$.
Thus, returning to the full induced density:
\begin{align*}
&p_{flickr}(s(x)) \cdot \int_{\hat{s}} p(\hat{s}\vert s)
\frac{p_{1}(\hat{s}(x))}{p_{flickr}(\hat{s}(x))} \\
&=
p_{flickr}(s(x)) \cdot \sum_{k=0}^n 
p(\hat{s}\vert s)
\frac{p_{1}(\hat{s}(x))}{p_{flickr}(\hat{s}(x))} \\
&= 
p_{flickr}(s(x)) \cdot \sum_{k=0}^n 
\left[{n \choose k} s^k (1-s)^{n-k}\right]
\frac{
    \left[\frac{{n \choose k} B(k+\alpha+1,n-k+\beta)}
    {B(\alpha+1,\beta)}\right]
}
{
    \left[\frac{{n \choose k} B(k+\alpha,n-k+\beta)}
    {B(\alpha,\beta)}\right]
} \\
&=
\left[
    \frac{s^{\alpha-1}\cdot (1-s)^{\beta-1}} { B(\alpha,\beta)} 
\right]\cdot \sum_{k=0}^n 
\left[{n \choose k} s^k (1-s)^{n-k}\right]
\frac{ B(k+\alpha+1,n-k+\beta) } { B(k+\alpha,n-k+\beta)} \cdot
\frac{ B(\alpha,\beta) } { B(\alpha+1,\beta) } \\
&=
\left[
    \frac{s^{\alpha-1}\cdot (1-s)^{\beta-1}} { B(\alpha+1,\beta)} 
\right]\cdot \sum_{k=0}^n 
\left[{n \choose k} s^k (1-s)^{n-k}\right]
\frac{ B(k+\alpha+1,n-k+\beta) } { B(k+\alpha,n-k+\beta)} 
\end{align*}

Now in general, note that
\begin{align*}
    \frac{B(x+1,y)}{B(x,y)} &=
    \frac{\frac{\Gamma(x+1)\Gamma(y)}{\Gamma(x+y+1)}}
    {\frac{\Gamma(x)\Gamma(y)}{\Gamma(x+y)}} = 
    \frac{\Gamma(x+1)}{\Gamma(x)}\cdot
    \frac{\Gamma(x+y)}{\Gamma(x+y+1)}
    = \frac{x}{x+y} &\text{for $x, y \in \mathbb{N}$}.
\end{align*}

Applying this identity to the above and continuing to simplify:
\begin{align*}
&= \left[
    \frac{s^{\alpha-1}\cdot (1-s)^{\beta-1}} { B(\alpha+1,\beta)} 
\right]\cdot \sum_{k=0}^n 
\left[{n \choose k} s^k (1-s)^{n-k}\right]
\frac{ k+\alpha } { n + \alpha + \beta } \\ 
&= \left[
    \frac{s^{\alpha-1}\cdot (1-s)^{\beta-1}} { B(\alpha+1,\beta)} 
\right]\cdot 
\mathbb{E}_{k\sim \text{Binomial}(n, s)}\left[ 
\frac{k+\alpha} {n+\alpha+\beta}\right]   \\
&= \left[
    \frac{s^{\alpha-1}\cdot (1-s)^{\beta-1}} { B(\alpha+1,\beta)} 
\right]\cdot \frac{n\cdot s + \alpha} {n+\alpha+\beta}   \\
&= \frac{n} {n+\alpha+\beta} \cdot 
\frac{s^{(\alpha+1)-1}\cdot (1-s)^{\beta-1}} { B(\alpha+1,\beta)} 
+ \frac{\alpha}{n+\alpha+\beta} 
\cdot \frac{s^{\alpha-1}\cdot (1-s)^{\beta-1}} { B(\alpha+1,\beta)} \\
&= \frac{n} {n+\alpha+\beta} \cdot 
\frac{s^{(\alpha+1)-1}\cdot (1-s)^{\beta-1}} { B(\alpha+1,\beta)} 
+ \frac{\alpha}{n+\alpha+\beta} \cdot 
\left(\frac{\alpha + \beta}{\alpha} \cdot \frac{B(\alpha+1,
\beta)}{B(\alpha, \beta)}\right) 
\cdot \frac{s^{\alpha-1}\cdot (1-s)^{\beta-1}} { B(\alpha+1,\beta)} \\
&= \frac{n} {n+\alpha+\beta} \cdot \text{Beta}(\alpha+1, \beta)
+ \frac{\alpha + \beta}{n+\alpha+\beta} \cdot \text{Beta}(\alpha, \beta)
\end{align*}
which matches precisely the result shown in Section~\ref{sec:id_bias}.

\clearpage

\section{Analysis of Original Data}
\label{app:orig_data}
In Section~\ref{sec:remeasure}, we remeasured selection frequencies. Here
we set out to verify the existence of the hypothesized bias in the original
collected data.

To this end, we reimplemented the sampling component of the algorithm exactly as described
in~\citep{recht2018imagenet} and the corresponding code release, using the
\texttt{pandas} Python package. The source code is available
in our code release, along with a serialized version of the data
collected by~\citet{recht2018imagenet}\footnote{\url{https://github.com/MadryLab/dataset-replication-analysis}}.
(As a sanity check, we verified that all of the results hold
using the exact code published by~\citet{recht2018imagenet}\footnote{Since we
only study the sampling component, we opt to rewrite a specialized analysis script
that has the benefit of being significantly shorter, simpler, and faster.})

\subsection{Sampled dataset accuracy increases with more workers}
\label{app:origdata_jackknife}
We begin by showing that the accuracy we observe on ImageNet-v2 depends on the
number of workers used to sample the dataset. 
We gradually decrease the number of workers $n$ used in computing observed
selection frequencies to study the effect of noise on statistic matching. We
find that model accuracy on the resulting replicated dataset degrades as $n$
decreases. For example, the accuracy gap from \va to the replication
increases from 12\% when $n = 10$, to 14\% when $n = 5$. This is consistent
with our model of statistic matching bias: fewer annotators means noisier
observed selection frequencies $\hs{x}$, which in turn amplifies the effect
of the bias, driving down model accuracies.

We use the
frequency-adjusted accuracy introduced in Section~\ref{sec:naive},
to estimate model accuracy on a version of the candidate pool reweighted to have
the same selection frequency distribution as ImageNet-v1:
\begin{align}
    \naiveestflickr = \sum_{k=0}^{n} 
        \evfl{\f(x_{flickr}) \bigg\vert \hs{x_{flickr}} = \frac{k}{n}} 
        \cdot p_1\left(\hs{x_1} =\frac{k}{n}\right).
\end{align}
This estimator is analogous to the ImageNet-v2 selection process
of~\citet{recht2018imagenet}, but operates by reweighting the candidate pool
rather than filtering it. 

\begin{figure}[hb]
    \centering
    \includegraphics[width=0.6\textwidth]{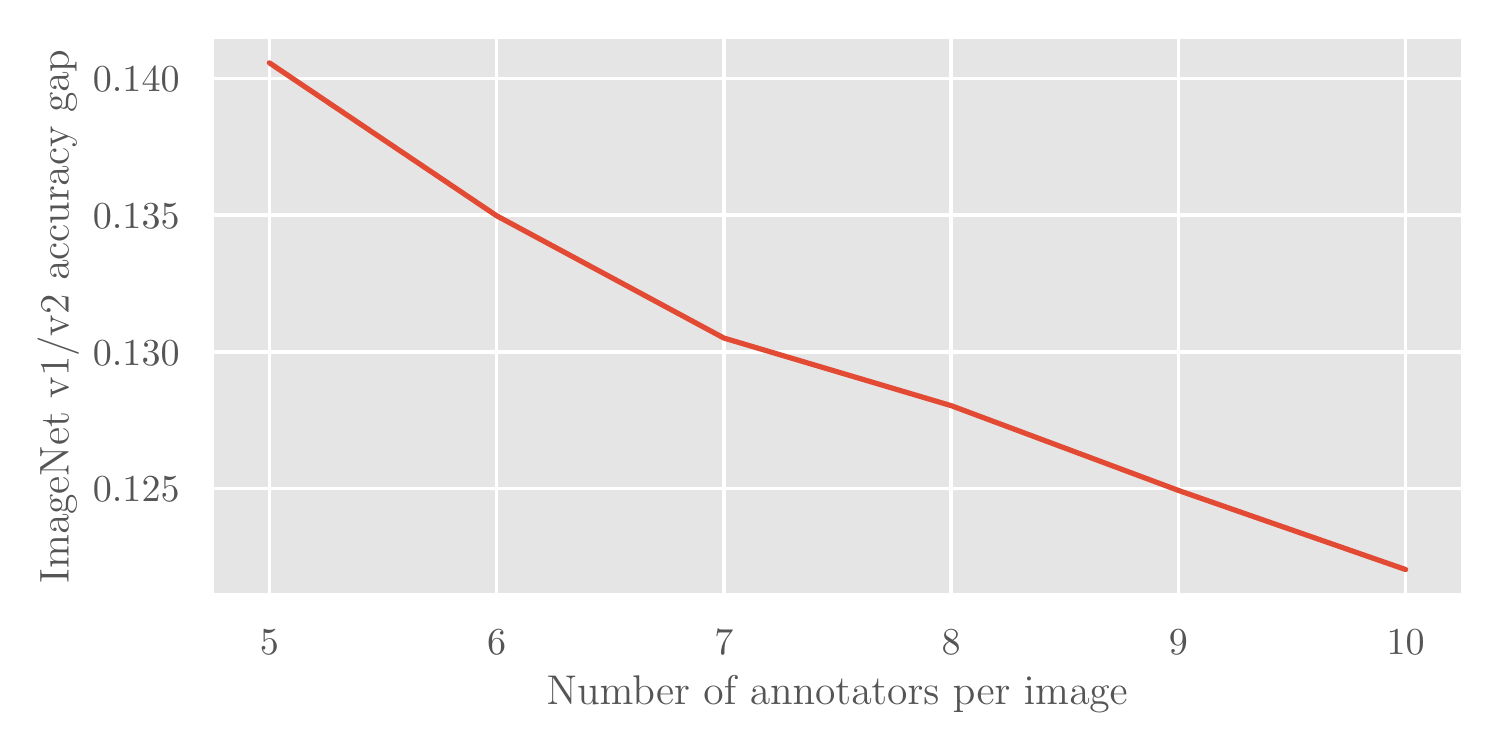}
    \caption{The frequency-adjusted accuracy gap between ImageNet-v1 and
    ImageNet-v2, using a varying number of annotators to estimate selection
    frequencies. The gap continually decreases, and does not
    plateau at 10 annotators. Bootstrapped 95\% confidence intervals are
    shown (shaded).} 
    \label{fig:orig_data_naive_est}
\end{figure}

We plot this estimator in Figure~\ref{fig:orig_data_naive_est}, varying $n$ from $5$ to $10$. We
find that the gap between the adjusted accuracy and ImageNet accuracy shrinks as
$n$ grows, until shrinking to (and not plateauing at) $12.3\%$ at $n = 10$. This
behavior is predicted by statistic matching bias, and suggests that in the
infinite-annotator limit the ImageNet-v2 accuracy is higher. (Ideally, we
could estimate the infinite-annotator limit using the data
of~\citet{recht2018imagenet}, but there are not enough [10] annotators in
the original data to get a reliable estimate.)

\subsection{Measuring the selection frequency gap using held-out workers}
\label{app:sfgap}
Recall that most images collected by \citet{recht2018imagenet} have 10 annotations
per image, and that to get an unbiased estimate of the selection frequency we
must ``hold out'' some annotations per image (e.g. we should not use the same
annotations for both matching and measuring selection frequencies).
To that end, we split the annotations for each candidate image into a ``train''
set and a ``test'' set. We then mimic the original \vb creation process using
only the train set of annotations, and use the remaining test annotations for
each image to obtain a held-out measurement of selection frequency. 

Since~\citet{recht2018imagenet} collect 10 annotations for most candidate
images, and since the original buckets used in the matching process are split
by boundaries $\{0, 0.2, 0.4, 0.6, 0.8, 1\}$, we use 5
annotations\footnote{We choose 5 annotations specifically since, as in the
10-annotation case, images fall into the same relative locations in each
bin---other choices of annotations per image are severely affected by binning
effects.} for each image in the train set and reserve the rest for
independently measuring the held-out frequency.

The results of this experiment are given in Table~\ref{tab:heldout_sf}. We
repeat the experiment using both 5 and 10 annotators to estimate ImageNet-v1
selection frequencies. In both cases, the
average selection frequency of the in-sample images overestimates the heldout
(true) selection frequency by 2-3\%, and the resulting replicated dataset has
lower selection frequency than ImageNet-v1.

\begin{table}[h]
    \centering
    \caption{Average in-sample and heldout selection frequencies for the
    experiment described in~\ref{app:sfgap}---the top (bottom) table presents the result
    of using 10 (5) annotators per image to estimate ImageNet-v1 selection
    frequencies. We use five annotators per image to 
    estimate selection frqeuencies, then use the filtering algorithm
    of~\citet{recht2018imagenet} to get a replicated dataset
    meant to match the selection frequency distribution of ImageNet-v1. 
    The results show that (a) bias results in the average selection frequency of the
    new sample being {\em lower} than that of ImageNet-v1, and that (b) the bias
    is undetectable without heldout samples.} 
    \vspace{1em}
    \label{tab:heldout_sf}
    \begin{tabular}{@{}lcc@{}} 
        \toprule
        & {\bf ImageNet-v1} & {\bf Sampled replication} \\ \midrule
        Average selection frequency & 0.71 & 0.71 \\
        Heldout selection frequency & 0.71 & 0.69 \\ \bottomrule
    \end{tabular}
    \begin{tabular}{@{}lcc@{}} 
        \toprule
        & {\bf ImageNet-v1} & {\bf Sampled replication} \\ \midrule
        Average selection frequency & 0.71 & 0.73 \\
        Heldout selection frequency & 0.71 & 0.70 \\ \bottomrule
    \end{tabular}
\end{table}

\paragraph{Effect size.}
The difference in held-out mean selection frequencies between \va and \vb
in the table above is smaller than the one we observe in the newly collected data. 
However, as discussed in Section~\ref{sec:remeasure}, the size
of the average gap is not necessarily predictive of the size of the accuracy
correction. 
The latter depends on the distributional difference between true \va
and \vb selection frequencies, rather than on just their first moments. 
In particular, observing different mean selection frequencies for \va and \vb is a
sufficient but not necessary condition for there to be an accuracy gap\footnote{As
a concrete example, suppose that 50\% of the annotators used were low-quality
and did not complete the task (i.e., selected no images)---this
would artificially shrink the mean selection frequency gap by 50\%, but would
not affect the accuracy adjustment.}.

Unfortunately, getting an accurate estimate of the gap on the original data
seems impossible: first, there are insufficient workers to reliably apply any of our
techniques from Section~\ref{sec:quantifying}. Furthermore, with $k$ held-out
workers, we can only estimate the first $k$ moments of the true selection
frequency distributions $p_i(s)$, even if we had infinitely many images.
Overall, the problem is thus largely underdetermined. 

We can, however, show that original data is plausibly consistent with the 4\%
accuracy gap estimated using the new data (i.e., that such a gap is not ruled
out). Specifically, in \ref{app:origdata_jackknife},
the original accuracy difference between Flickr and ImageNet-v1 was 20.8\%. When
using five (ten) annotators per image, this gap shrunk to 14.0\% (12.2\%). 
In the newly collected data, the gaps for
the original distributions, 5-annotator adjustment, and 10-annotator adjustment are
11.7\%, 8.5\%, and 7.3\% respectively (again, by using the same reweighting
scheme as Appendix \ref{app:origdata_jackknife}). 
Therefore, accuracy adjustments incurred by using more
workers on the original data are significantly (about two times) larger than the
corresponding accuracy adjustments on the newly collected data, and so we expect
to see a larger total correction than our estimated 8.1\% correction.

\subsection{Source selection frequencies determine sampled dataset accuracy}
We next explore how the choice of source distribution impacts the resulting
sampled dataset. We use a setup similar to that of the last experiment, in which
we use five workers for the selection process. Then, using four hold-out samples
from each image, we create a new candidate data pool called Flickr-E
(Flickr-Easy) by including only the images which at least two out of the
four heldout workers selected.

We then perform 5-worker statistic matching, both from Flickr and from Flickr-E
to ImageNet-v1. In the absence of bias, the source distribution should not
affect the accuracy of the resulting classifier. In contrast, we find that the
dataset replication obtained from Flickr-v2 has comparable (within 0.2\%)
average selection frequency, but significantly higher accuracy (by
\textasciitilde 3\%) than the
replication obtained from the unfiltered candidate pool (62\%).
This discrepancy further corroborates the hypothesis that ImageNet-v2 accuracies
are impacted by the statistical bias that we identify in this work.

\clearpage

\section{Non-parametric Adjusted Accuracy Estimation}
\label{app:kde}
In Section~\ref{sec:quantifying} we explore various methods of estimating the
adjusted accuracy $\adjacc$ from the observable $\hat{s}$ samples.
Section~\ref{sec:naive} presents the {\em na\"ive estimator}, computed by using
$\hat{s}$ directly in place of $s$ in the formula for $\adjacc$. In
Section~\ref{sec:jackknife}, we show that using the statistical jackknife, we
can estimate and account for the bias in the na\"ive estimator to better
estimate the true adjusted accuracy. Here, we first justify the application of
the jackknife in Section~\ref{sec:jackknife}, by proving the consistency of the
estimator and that bias is roughly linear in (and in fact underestimated by)
$1/n$.

\subsection{Justifying the use of the statistical jackknife}
\label{app:jk_justification}
Recall that in Section~\ref{sec:quantifying}, the quantity of interest is the
following {\em adjusted accuracy}:
\begin{equation}
    \adjacc = \int_s \mathbb{E}_{\dvb}[f(x)|s(x)=s] \cdot p_1(s)\ ds.
\end{equation}
In Section~\ref{sec:naive}, we introduced the following na\"ive estimator
\begin{equation}
    \naiveest = \sum_{k=0}^{n} \evb{\f(x_2)\vert \hs{x_2} = \frac{k}{n}} \cdot p_1\left(\hs{x_1} =\frac{k}{n}\right),
\end{equation}
which evaluates to $\adjacc$ if $\hs{x} = \s{x}$, and assuming the expectations
above can be computed exactly.

\paragraph{Consistency of the na\"ive estimator.} In order for our application of
the statistical jackknife to be valid, we must show that the na\"ive estimator
is a {\em consistent} estimator of $\adjacc$---that is, as $n \rightarrow
\infty$, $\naiveest \rightarrow \adjacc$. Note that since we operate in the
regime where the number of distinct images $m$ greatly exceeds the number of
annotators per image $n$, we will assume that the expectations above can be
computed exactly. Note that the estimator remains consistent even if this is not
the case, with the additional constraints that $m \rightarrow \infty$ and $m / n
\rightarrow \infty$, but this greatly complicates the proof and we will show
empirically that the estimator is robust to changes in $m$ in the relevant
regime.

Note that in the ``infinite $m$'' regime, the variance of the na\"ive 
estimator is $0$. Thus, all of the error is due to bias in the estimator. In the
following, we assume that $p_1(s)$, $p_2(s)$, and $p_2(s|f=1)$ are continuous
differentiable densities bounded away from zero and with bounded derivatives
($|d^r/dx^r\ p_i(x)| < C$).
\begin{align}
    p_i\left(\hs{x}=\frac{k}{n}\right) &= \int_0^1 p_i(s)\cdot \binomial{n,k,s}\ ds \\
    &= \int_0^1 p_i(s)\cdot {n \choose k} \cdot s^k \cdot (1-s)^{n-k}\ ds \\
    &= \int_0^1 p_i(s)\cdot \frac{s^k (1-s)^{n-k}}{(n+1)\cdot B(k+1, n-k+1)}\ ds &B(\cdot,\cdot)\text{ is the Euler beta function}\\
    &= \frac{1}{n+1}\int_0^1 p_i(s)\cdot \text{Beta}(s;k+1,n-k+1)\ ds \\
    &= \frac{1}{n+1}\mathbb{E}_{s\sim \text{Beta}(\cdot;k+1,n-k+1)}[p_i(s)]
\end{align}
Using a Taylor expansion of $p_i(s)$ around
$\mathbb{E}_{s\sim\text{Beta}(\cdot;k+1,n-k+1)}[s]$, we can bound the above
expression:
\begin{align*}
    &= \frac{1}{n+1}\mathbb{E}\left[
        p_i(\mathbb{E}[s]) 
        + (s-\mathbb{E}[s])p_i'(E[s]) + \frac{(s-\mathbb{E}[s])^2}{2}p_i''(\mathbb{E}[s])
        + \sum_{r=3}^\infty \frac{(s - \mathbb{E}[s])^r}{r!} p_i^{(r)}(\mathbb{E}[s])
    \right]  \\
    &= \frac{1}{n+1}\left(p_i(\mathbb{E}[s]) + \frac{1}{2}\text{Var}[s]\cdot p_i''(\mathbb{E}[s]) + O\left(\frac{1}{n^{7/2}}\right) \right) \\
    &= \frac{1}{n+1}p_i\left(\frac{k+1}{n+2}\right) + \frac{(k+1)(n-k+1)}{(n+1)(n+2)^2(n+3)}\cdot p_i''(\mathbb{E}[s]) + O\left(\frac{1}{n^{7/2}}\right) 
\end{align*}
Now, using the presumed boundedness of derivatives we can write:
\begin{align}
    \label{eq:term_bound}
    \left| p_i\left(\hs{x}=\frac{k}{n}\right) - 
    \frac{1}{n+1}\cdot p_i\left(\frac{k+1}{n+2}\right) \right| 
    &\leq \frac{(k+1)(n-k+1)}{(n+1)(n+2)^2(n+3)} \cdot p_i''(\mathbb{E}[s]) 
    + O\left(\frac{1}{n^{7/2}}\right) 
\end{align}

\begin{align*}
    |\adjacc - \naiveest| &= p_2(f(x) = 1)\left|
        \int_0^1 \frac{p_2(s|f(x) = 1)}{p_2(s)} p_1(s)\ ds
        - \sum_{k = 0}^{n} \frac{
            p_2(\hs{x} = \frac{k}{n} | f(x) = 1)
        }{
            p_2(\hs{x} = \frac{k}{n})
        } p_1(\hs{x} = \frac{k}{n})
    \right| \\
    &\leq \left|\int_0^1 \frac{p_2(s|f(x)=1)}{p_2(s)}p_1(s)\ ds 
    -  \frac{1}{n+1}\sum_{k=0}^{n} \frac{
        p_2\left(\frac{k+1}{n+2}\big| f(x) = 1\right)
        }{ p_2\left(\frac{k+1}{n+2}\right)} 
        p_1\left(\frac{k+1}{n+2}\right)  \right| \\
    &+  \left| \frac{1}{n+1}\sum_{k=0}^{n} \frac{
        p_2\left(\frac{k+1}{n+2}\big| f(x) = 1\right)
        }{ p_2\left(\frac{k+1}{n+2}\right)} 
        p_1\left(\frac{k+1}{n+2}\right)  
    - \frac{
        p_2\left(\hs{x} = \frac{k}{n}\big| f(x) = 1\right)
        }{ p_2\left(\hs{x} = \frac{k}{n}\right)} 
        p_1\left(\hs{x} = \frac{k}{n}\right)  \right| 
\end{align*}
Now, note that the first term in the above is simply the error in the Riemmann
sum approximation of the integral, which vanishes as $n \rightarrow \infty$. The
second term is bounded by $n$ times the error in each individual term of the
sum, which we bounded as $O(n^{-2})$ in Equation~\eqref{eq:term_bound}.

\begin{figure}[!h]
	\centering
   \includegraphics[width=0.8\textwidth]{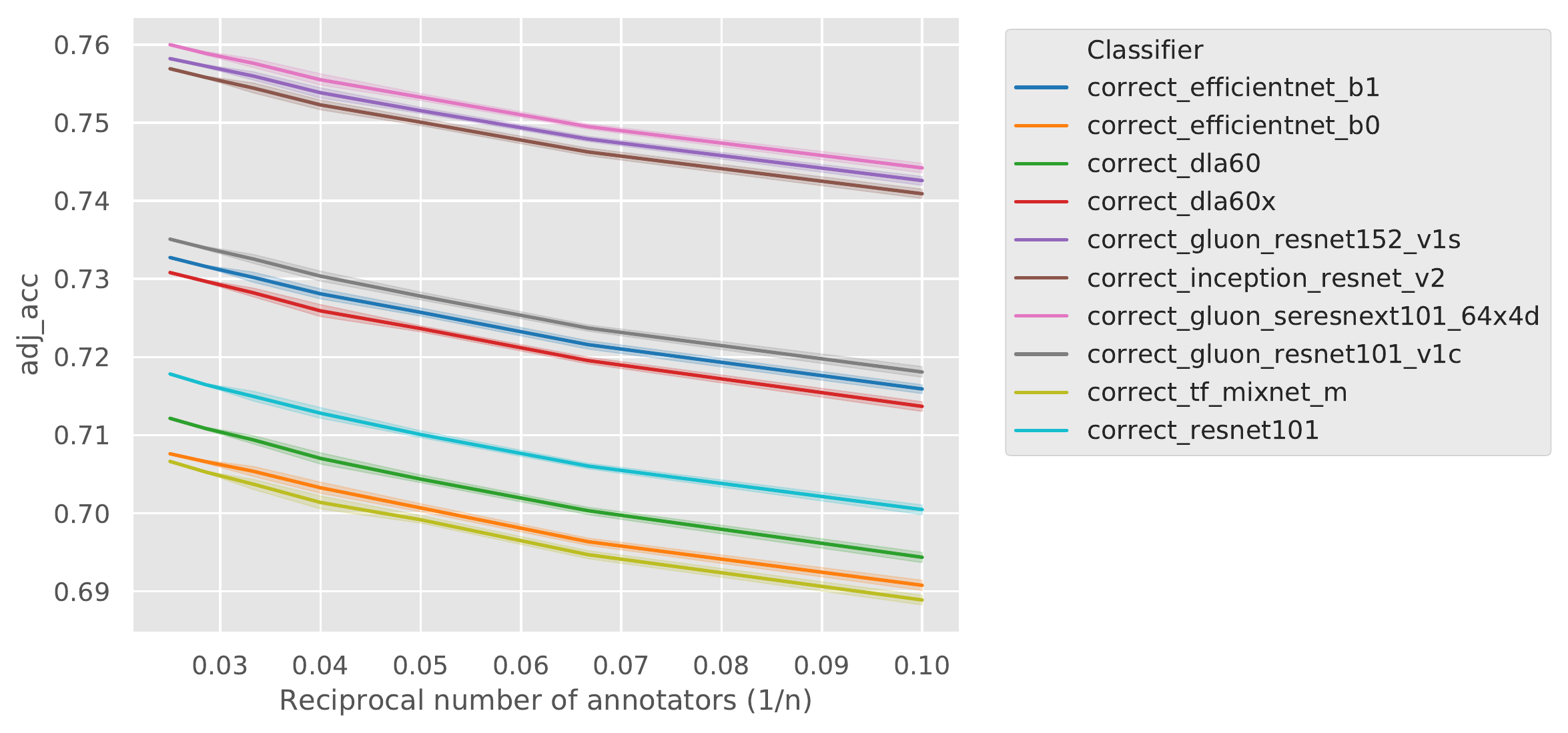}
 	\caption{We plot the jackknife adjusted accuracy estimators of 10 models. On
 the y axis is shown the value of the $n$-sample jackknife estimate, with $1/n$
 on the x axis. The fact that the plot is nearly linear suggests that our bias is
     indeed dominated by a $O(1/n)$ term, thus further justifying our use
 of the statistical jackknife in Section~\ref{sec:jackknife}. Furthermore, the
 slightly accelerating slope as one moves left on the plot indicates that any
 error in the jackknife estimate is likely to be an underestimation of bias,
 rather than an overestimation.}
 	\label{fig:analytic}
\end{figure}

\paragraph{Near-linearity of bias and likely underestimation.}
Recall that for the statistical jackknife to yield a reliable estimate of the adjusted
accuracy, the bias in the na\"ive estimator must be analytic in
$\frac{1}{n}$, and in particular should be dominated by a $O(\frac{1}{n})$
term (as this corresponds to precisely the term accounted for by the
jackknife). In Figure~\ref{fig:analytic} we show that the bias estimated by
our jackknife procedure is indeed roughly linear in $1/n$, but grows slightly
faster than $\frac{1}{n}$, likely leading the jackknife to provide an
underestimate. 

\clearpage

\section{Model Fitting}
\label{app:model_fitting}
In this section, we describe our methods for parametric modeling.

\subsection{Confidence Intervals}
To construct 95\% confidence intervals we perform 450 bootstrapped estimates (over
the included images) of adjusted accuracy for each classifier. We then plot the
2.5\% and 97.5\% percentiles from the bootstrap estimates as the confidence
intervals for each classifier.

\subsection{Varying Annotators}
\begin{figure}[!ht]
	\centering
  \input{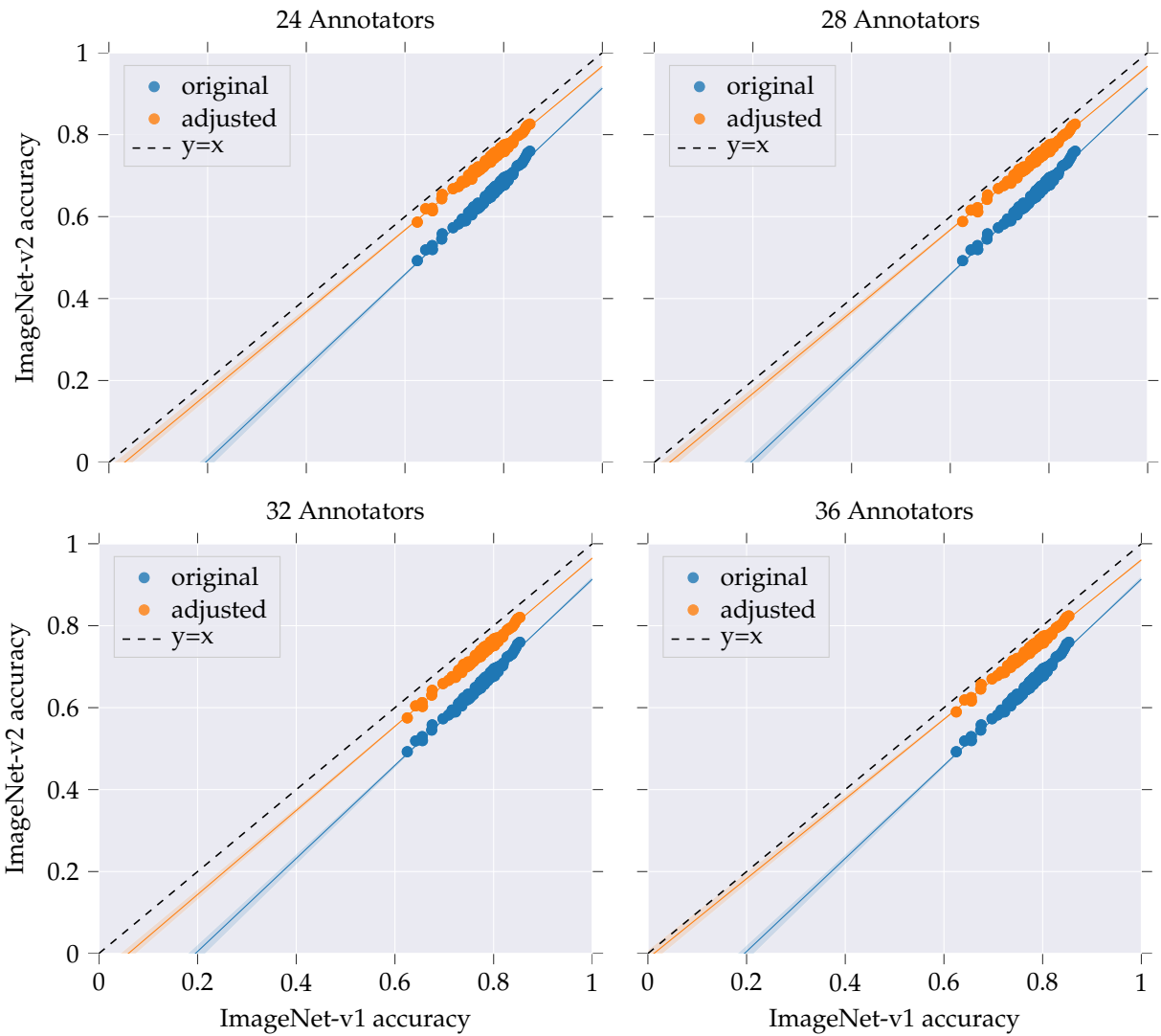}
	\caption{Replicating the \va vs \vb accuracy plot using different numbers of
    annotators. We obtain similar results even as we decrease the number of
    annotators by less than half.}
	\label{fig:vary_an}
\end{figure}

We plot the results of varying the number of annotators in
Figure~\ref{fig:vary_an}.

\subsection{Varying Model Expressiveness}
\begin{figure}[!ht]
	\centering
    \input{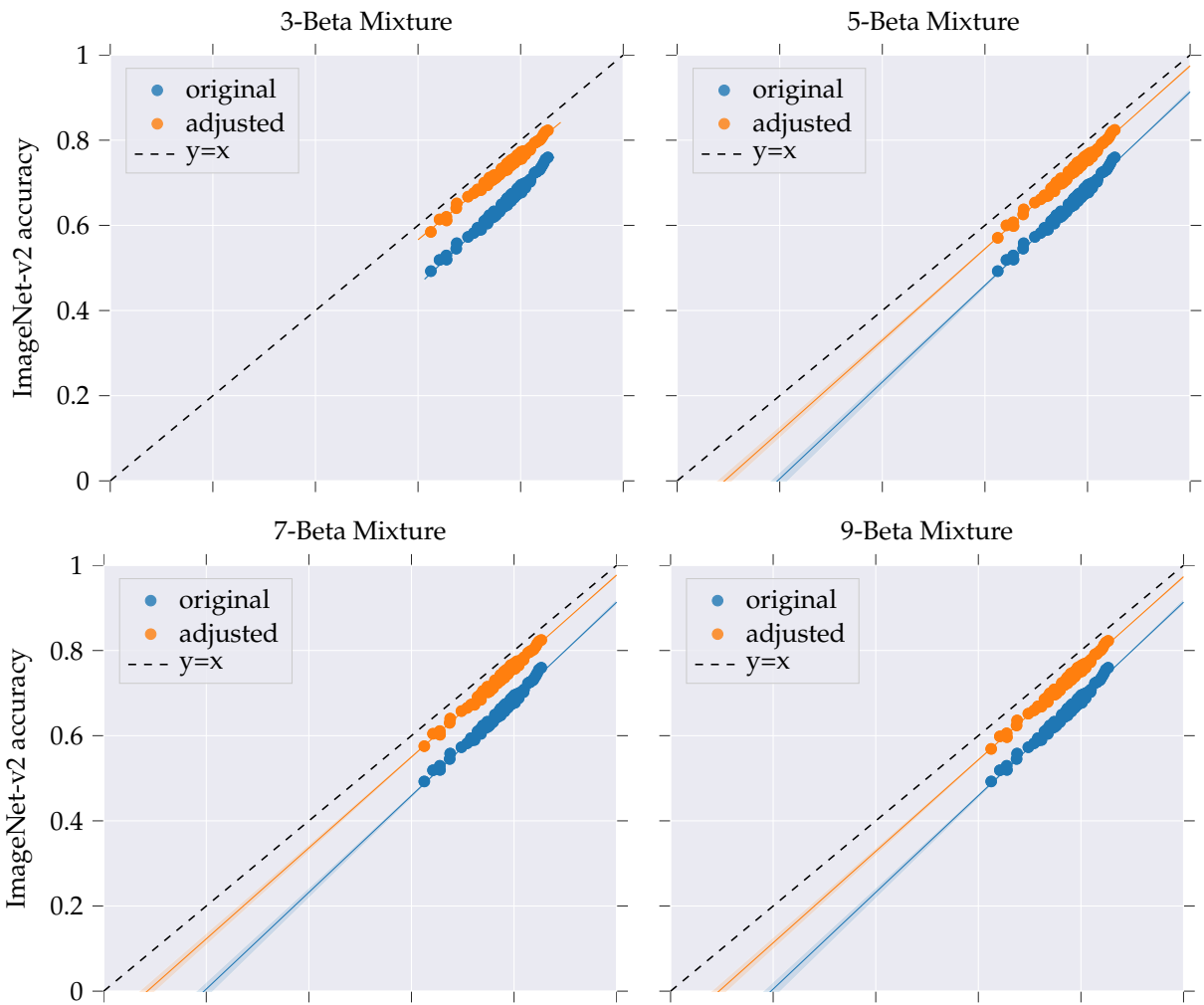}
	\caption{Replicating the \va vs \vb accuracy plot using different numbers of
    parameters. We obtain similar results even as we increase the number of
    parameters by more than four-fold.}
	\label{fig:vary_mix}
\end{figure}

We plot the results of varying the number of parameters (here, by changing the
number of beta distributions in our mixture) in Figure~\ref{fig:vary_mix}.

\subsection{EM Algorithm for Mixture Fitting}
To fit the parameters of the beta-binomial mixture model we apply the
Expectation-Maximization algorithm, optimizing over mixture coefficients
$\{\pi_i\}$, as well as parameters of each mixture element
$\{(\alpha_i, \beta_i)\}$. Our application of the EM algorithm is rather
canonical---first, we compute membership probabilities $p_{j}^{i}$ for each
example $j$ with respect to each mixture element $i$, then minimize the
weighted log-likelihood with respect to the mixture probabilities.
Pseudocode is given in Algorithm~\ref{alg:em}.

\begin{algorithm}
    \caption{Our instantiation of the EM algorithm}
    \label{alg:em}
    \begin{algorithmic}
    \STATE {\bfseries Input:} A set of size $n$ of empirical selection
    probabilities
    $\{\hat{s}_j\}$, the observed rate at which each image was selected,
    number of mixture components $k$.
    \STATE Start with random guesses for all parameters:
    $$\alpha_i, \beta_i, \pi_i \gets \text{random}$$
    \FOR{each training iteration}
    \STATE {\bf 1.} Calculate membership probabilities for each observed
    element:
    $$p_{j}^i = \frac{\pi_i \cdot p(\hat{s};\alpha_i,\beta_i,40)}
    {\sum_{r=1}^k \pi_r \cdot p(\hat{s};\alpha_r,\beta_r,40)} 
    \qquad\forall\ j
    \in [n], i \in [k],$$
    where $p(\cdot;\alpha,\beta,40)$ is likelihood under the beta-binomial
    distribution with $40$ samples.
    \STATE {\bf 2.} As is standard in EM, update the parameters by
    minimizing the expected log-likelihood, weighted by the membership
    probabilities---i.e. update
    $$\alpha_i, \beta_i = \min_{\alpha,\beta} \sum_{j=1}^n p_j^i \cdot
    \ell(\hat{s}_j;\alpha_i,\beta_i,40),$$
    where $\ell(\cdot) = -\log(p(\cdot))$ is the negative log-likelihood, and 
    $$\pi_i = \frac{\sum_{j=1}^n p_j^{i}} {\sum_{r=1}^k \sum_{j=1}^n
    p_j^{r}}.$$
    \ENDFOR
\end{algorithmic}

\end{algorithm}

\clearpage

\section{Full Model Results}
\label{app:fullmodel}
In Appendix Table~\ref{tab:models}, we detail the set of models we use in our evaluation 
along with their corresponding Top-1 accuracies on ImageNet-v1 and -v2 validation sets. 
We use 
open-source pre-trained implementations from 
\url{https://github.com/rwightman/pytorch-image-models} for all
architectures. 

\begin{longtable}{lccc}
\toprule
\textbf{Model} & \phantom{x} & \textbf{v1} & \textbf{v2} \\
\midrule
\texttt{tf\_mobilenetv3\_small\_minimal\_100} && 63.070\% & 48.270\% \\ 
\texttt{dla46\_c} && 64.950\% & 51.330\% \\ 
\texttt{tf\_mobilenetv3\_small\_075} && 65.490\% & 50.800\% \\ 
\texttt{dla46x\_c} && 66.130\% & 52.200\% \\ 
\texttt{tf\_mobilenetv3\_small\_100} && 67.500\% & 53.960\% \\ 
\texttt{dla60x\_c} && 68.170\% & 55.660\% \\ 
\texttt{resnet18} && 70.420\% & 56.850\% \\ 
\texttt{gluon\_resnet18\_v1b} && 71.280\% & 57.610\% \\ 
\texttt{seresnet18} && 71.840\% & 58.200\% \\ 
\texttt{tf\_mobilenetv3\_large\_minimal\_100} && 71.910\% & 57.870\% \\ 
\texttt{hrnet\_w18\_small} && 72.860\% & 58.120\% \\ 
\texttt{tv\_resnet34} && 73.080\% & 60.060\% \\ 
\texttt{spnasnet\_100} && 73.760\% & 61.040\% \\ 
\texttt{tf\_mobilenetv3\_large\_075} && 73.850\% & 59.430\% \\ 
\texttt{gluon\_resnet34\_v1b} && 74.470\% & 61.630\% \\ 
\texttt{mnasnet\_100} && 74.620\% & 61.020\% \\ 
\texttt{densenet121} && 74.650\% & 61.810\% \\ 
\texttt{dla34} && 74.680\% & 61.510\% \\ 
\texttt{seresnet34} && 74.820\% & 62.330\% \\ 
\texttt{resnet34} && 74.990\% & 62.240\% \\ 
\texttt{hrnet\_w18\_small\_v2} && 75.000\% & 61.540\% \\ 
\texttt{fbnetc\_100} && 75.080\% & 61.240\% \\ 
\texttt{resnet26} && 75.270\% & 62.730\% \\ 
\texttt{semnasnet\_100} && 75.690\% & 62.360\% \\ 
\texttt{tf\_mobilenetv3\_large\_100} && 75.710\% & 61.400\% \\ 
\texttt{mobilenetv3\_rw} && 75.740\% & 61.870\% \\ 
\texttt{tv\_resnet50} && 75.820\% & 62.600\% \\ 
\texttt{dpn68} && 76.020\% & 63.000\% \\ 
\texttt{tf\_mixnet\_s} && 76.210\% & 62.040\% \\ 
\texttt{tf\_efficientnet\_b0} && 76.240\% & 63.050\% \\ 
\texttt{densenet169} && 76.370\% & 63.450\% \\ 
\texttt{hrnet\_w18} && 76.500\% & 64.560\% \\ 
\texttt{mixnet\_s} && 76.570\% & 62.840\% \\ 
\texttt{dla60} && 76.800\% & 64.610\% \\ 
\texttt{efficientnet\_b0} && 76.820\% & 64.050\% \\ 
\texttt{seresnext26\_32x4d} && 76.980\% & 64.050\% \\ 
\texttt{resnet26d} && 77.020\% & 63.970\% \\ 
\texttt{resnet101} && 77.090\% & 65.020\% \\ 
\texttt{tf\_mixnet\_m} && 77.120\% & 63.540\% \\ 
\texttt{tf\_efficientnet\_b0\_ap} && 77.130\% & 64.290\% \\ 
\texttt{tf\_efficientnet\_cc\_b0\_4e} && 77.190\% & 64.110\% \\ 
\texttt{tf\_efficientnet\_es} && 77.200\% & 64.360\% \\ 
\texttt{inception\_v3} && 77.240\% & 65.090\% \\ 
\texttt{densenet161} && 77.240\% & 64.790\% \\ 
\texttt{densenet201} && 77.280\% & 64.480\% \\ 
\texttt{res2net50\_48w\_2s} && 77.420\% & 64.260\% \\ 
\texttt{gluon\_resnet50\_v1b} && 77.530\% & 65.130\% \\ 
\texttt{adv\_inception\_v3} && 77.680\% & 65.380\% \\ 
\texttt{mixnet\_m} && 77.710\% & 64.090\% \\ 
\texttt{gluon\_resnet50\_v1c} && 77.710\% & 65.180\% \\ 
\texttt{dpn68b} && 77.720\% & 64.830\% \\ 
\texttt{tf\_efficientnet\_cc\_b0\_8e} && 77.740\% & 64.410\% \\ 
\texttt{resnet152} && 77.760\% & 66.410\% \\ 
\texttt{tv\_resnext50\_32x4d} && 77.790\% & 65.130\% \\ 
\texttt{dla60\_res2next} && 77.980\% & 65.820\% \\ 
\texttt{hrnet\_w30} && 78.010\% & 65.860\% \\ 
\texttt{seresnet50} && 78.020\% & 65.160\% \\ 
\texttt{res2net50\_26w\_4s} && 78.050\% & 64.590\% \\ 
\texttt{seresnet101} && 78.070\% & 66.150\% \\ 
\texttt{dla60x} && 78.160\% & 66.090\% \\ 
\texttt{res2next50} && 78.180\% & 65.370\% \\ 
\texttt{tf\_inception\_v3} && 78.220\% & 65.480\% \\ 
\texttt{dla102} && 78.290\% & 65.710\% \\ 
\texttt{hrnet\_w44} && 78.300\% & 67.130\% \\ 
\texttt{dla169} && 78.380\% & 66.450\% \\ 
\texttt{wide\_resnet101\_2} && 78.430\% & 65.460\% \\ 
\texttt{wide\_resnet50\_2} && 78.430\% & 65.750\% \\ 
\texttt{tf\_efficientnet\_b1} && 78.530\% & 65.620\% \\ 
\texttt{res2net50\_14w\_8s} && 78.540\% & 65.180\% \\ 
\texttt{hrnet\_w32} && 78.600\% & 65.620\% \\ 
\texttt{dla60\_res2net} && 78.610\% & 65.550\% \\ 
\texttt{tf\_mixnet\_l} && 78.610\% & 65.750\% \\ 
\texttt{hrnet\_w40} && 78.670\% & 66.600\% \\ 
\texttt{efficientnet\_b1} && 78.690\% & 66.300\% \\ 
\texttt{tf\_efficientnet\_em} && 78.710\% & 65.560\% \\ 
\texttt{resnext50\_32x4d} && 78.790\% & 66.530\% \\ 
\texttt{dla102x} && 78.810\% & 66.140\% \\ 
\texttt{seresnet152} && 78.850\% & 66.540\% \\ 
\texttt{hrnet\_w48} && 78.860\% & 66.320\% \\ 
\texttt{gluon\_inception\_v3} && 78.880\% & 66.110\% \\ 
\texttt{res2net50\_26w\_6s} && 78.890\% & 66.200\% \\ 
\texttt{mixnet\_l} && 78.890\% & 66.180\% \\ 
\texttt{resnet50} && 79.000\% & 65.770\% \\ 
\texttt{hrnet\_w64} && 79.090\% & 67.650\% \\ 
\texttt{res2net50\_26w\_8s} && 79.100\% & 66.710\% \\ 
\texttt{xception} && 79.110\% & 66.320\% \\ 
\texttt{gluon\_resnet101\_v1b} && 79.110\% & 66.300\% \\ 
\texttt{gluon\_resnet50\_v1s} && 79.140\% & 66.220\% \\ 
\texttt{gluon\_resnet50\_v1d} && 79.200\% & 66.740\% \\ 
\texttt{tf\_efficientnet\_b1\_ap} && 79.330\% & 66.290\% \\ 
\texttt{tf\_efficientnet\_cc\_b1\_8e} && 79.360\% & 65.890\% \\ 
\texttt{dla102x2} && 79.400\% & 67.830\% \\ 
\texttt{seresnext50\_32x4d} && 79.420\% & 66.790\% \\ 
\texttt{resnext101\_32x8d} && 79.490\% & 66.660\% \\ 
\texttt{gluon\_resnext50\_32x4d} && 79.630\% & 67.610\% \\ 
\texttt{gluon\_resnet101\_v1c} && 79.660\% & 66.870\% \\ 
\texttt{resnext50d\_32x4d} && 79.700\% & 67.700\% \\ 
\texttt{tf\_efficientnet\_b2} && 79.730\% & 67.320\% \\ 
\texttt{res2net101\_26w\_4s} && 79.740\% & 66.750\% \\ 
\texttt{dpn98} && 79.830\% & 67.550\% \\ 
\texttt{dpn107} && 79.950\% & 67.490\% \\ 
\texttt{dpn131} && 80.020\% & 67.580\% \\ 
\texttt{gluon\_resnet152\_v1b} && 80.030\% & 67.610\% \\ 
\texttt{gluon\_xception65} && 80.070\% & 68.000\% \\ 
\texttt{ens\_adv\_inception\_resnet\_v2} && 80.080\% & 68.630\% \\ 
\texttt{efficientnet\_b2} && 80.080\% & 67.490\% \\ 
\texttt{gluon\_seresnext50\_32x4d} && 80.100\% & 67.800\% \\ 
\texttt{inception\_v4} && 80.170\% & 68.490\% \\ 
\texttt{gluon\_resnet152\_v1c} && 80.230\% & 67.660\% \\ 
\texttt{gluon\_resnet101\_v1s} && 80.320\% & 68.020\% \\ 
\texttt{mixnet\_xl} && 80.340\% & 68.000\% \\ 
\texttt{tf\_efficientnet\_el} && 80.550\% & 67.190\% \\ 
\texttt{seresnext101\_32x4d} && 80.570\% & 69.050\% \\ 
\texttt{gluon\_resnext101\_32x4d} && 80.580\% & 67.510\% \\ 
\texttt{dpn92} && 80.600\% & 66.750\% \\ 
\texttt{tf\_efficientnet\_b2\_ap} && 80.680\% & 67.380\% \\ 
\texttt{inception\_resnet\_v2} && 80.840\% & 68.750\% \\ 
\texttt{gluon\_resnext101\_64x4d} && 80.860\% & 69.050\% \\ 
\texttt{gluon\_resnet152\_v1d} && 80.870\% & 68.730\% \\ 
\texttt{gluon\_resnet101\_v1d} && 81.020\% & 67.960\% \\ 
\texttt{gluon\_seresnext101\_32x4d} && 81.060\% & 68.890\% \\ 
\texttt{gluon\_resnet152\_v1s} && 81.470\% & 68.980\% \\ 
\texttt{gluon\_seresnext101\_64x4d} && 81.700\% & 69.040\% \\ 
\texttt{tf\_efficientnet\_b3} && 81.810\% & 69.360\% \\ 
\texttt{gluon\_senet154} && 81.900\% & 69.930\% \\ 
\texttt{senet154} && 82.100\% & 70.020\% \\ 
\texttt{tf\_efficientnet\_b3\_ap} && 82.130\% & 69.920\% \\ 
\texttt{nasnetalarge} && 82.780\% & 71.660\% \\ 
\texttt{pnasnet5large} && 83.210\% & 71.970\% \\ 
\texttt{tf\_efficientnet\_b4} && 83.350\% & 71.920\% \\ 
\texttt{tf\_efficientnet\_b5} && 84.030\% & 72.200\% \\ 
\texttt{tf\_efficientnet\_b4\_ap} && 84.210\% & 72.130\% \\ 
\texttt{tf\_efficientnet\_b5\_ap} && 84.260\% & 73.520\% \\ 
\texttt{tf\_efficientnet\_b6} && 84.510\% & 72.940\% \\ 
\texttt{tf\_efficientnet\_b6\_ap} && 85.000\% & 74.570\% \\ 
\texttt{tf\_efficientnet\_b7\_ap} && 85.460\% & 75.110\% \\ 
			\bottomrule
\caption{Models used in our analysis with the corresponding 
	Top-1 on the ImageNet v1 and v2 validation sets.}
\label{tab:models}
\end{longtable}

\end{document}